\documentclass{article}

\usepackage[preprint]{main}

\usepackage[utf8]{inputenc} 
\usepackage[T1]{fontenc}    
\usepackage{hyperref}       
\usepackage{url}            
\usepackage{booktabs}       
\usepackage{amsfonts}       
\usepackage{nicefrac}       
\usepackage{microtype}      
\usepackage{xcolor}         
\usepackage{multirow} 
\usepackage{xcolor}
\usepackage{mdframed}
\usepackage{colortbl}
\usepackage{graphicx}
\usepackage{amsmath}
\usepackage{subcaption}
\graphicspath{{src/}}

\title{4D-Animal: Freely Reconstructing Animatable 3D Animals from Videos}

\author{
  Shanshan Zhong$^{1,2,}$\thanks{Work done while visiting at JHU.} , Jiawei Peng$^{1}$, Zehan Zheng$^{1}$, Zhongzhan Huang$^{2}$, Wufei Ma$^{1}$, \\
  \textbf{Guofeng Zhang$^{1}$, Qihao Liu$^{1}$, Alan Yuille$^{1}$, Jieneng Chen$^{1,}$\thanks{Corresponding author}} \\
  $^{1}$Johns Hopkins University \quad $^{2}$Sun Yat-sen University
}

\begin{document}

\maketitle

\begin{figure}[h]
    \centering
    \includegraphics[width=0.99\textwidth]{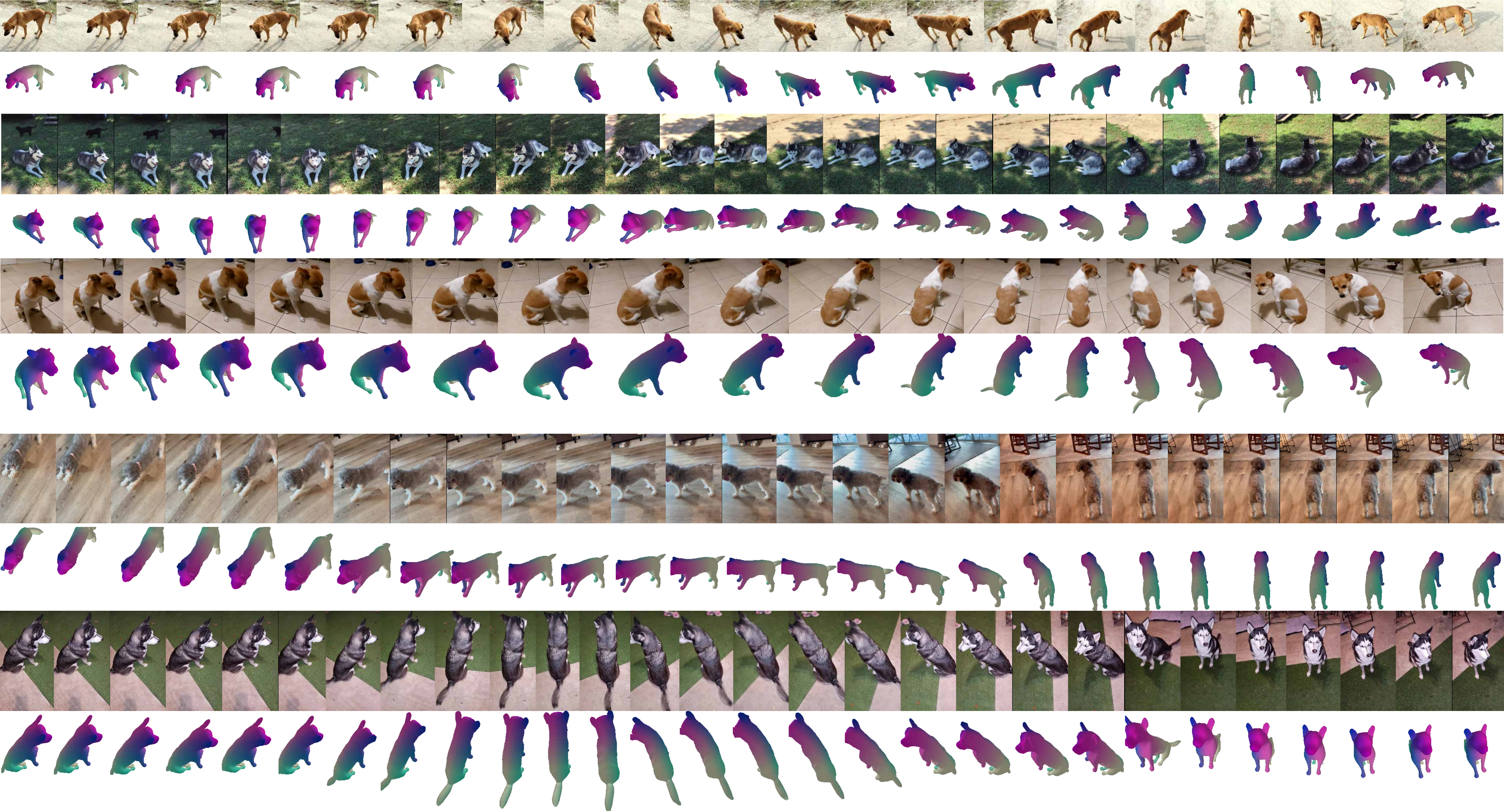}
    \caption{Visualization of 4D reconstruction results generated by the proposed 4D-Animal. Five examples of diverse in-the-wild videos are shown. } 
    \label{fig:example1}
\end{figure}%

\begin{abstract}
Existing methods for reconstructing animatable 3D animals from videos typically rely on sparse semantic keypoints to fit parametric models. However, obtaining such keypoints is labor-intensive, and keypoint detectors trained on limited animal data are often unreliable. To address this, we propose 4D-Animal, a novel framework that reconstructs animatable 3D animals from videos without requiring sparse keypoint annotations. Our approach introduces a dense feature network that maps 2D representations to SMAL parameters, enhancing both the efficiency and stability of the fitting process. Furthermore, we develop a hierarchical alignment strategy that integrates silhouette, part-level, pixel-level, and temporal cues from pre-trained 2D visual models to produce accurate and temporally coherent reconstructions across frames. Extensive experiments demonstrate that 4D-Animal outperforms both model-based and model-free baselines. Moreover, the high-quality 3D assets generated by our method can benefit other 3D tasks, underscoring its potential for large-scale applications. The code is released at \href{https://github.com/zhongshsh/4D-Animal}{\texttt{https://github.com/zhongshsh/4D-Animal}}.
\end{abstract}

\section{Introduction}
\label{sec:intro}

Reconstructing animatable 3D animals from videos has broad applications such as augmented and virtual reality~\cite{sabathier2024animal,kasten2024fast,das2024neural,yang2023reconstructing,yang2022banmo}. Among them, quadruped reconstruction is particularly important due to its relevance to human life. While it may seem similar to human reconstruction, the underlying technical difficulties differ significantly~\cite{li2024fauna,yang2023ppr}. Human models benefit from abundant 3D scans and motion capture data, enabling expressive models like SMPL and GHUM~\cite{loper2023smpl, xu2020ghum}. In contrast, the scarcity of 3D animal data makes their reconstruction significantly more difficult.

\begin{figure}[b]
  \centering
\includegraphics[width=0.89\linewidth]{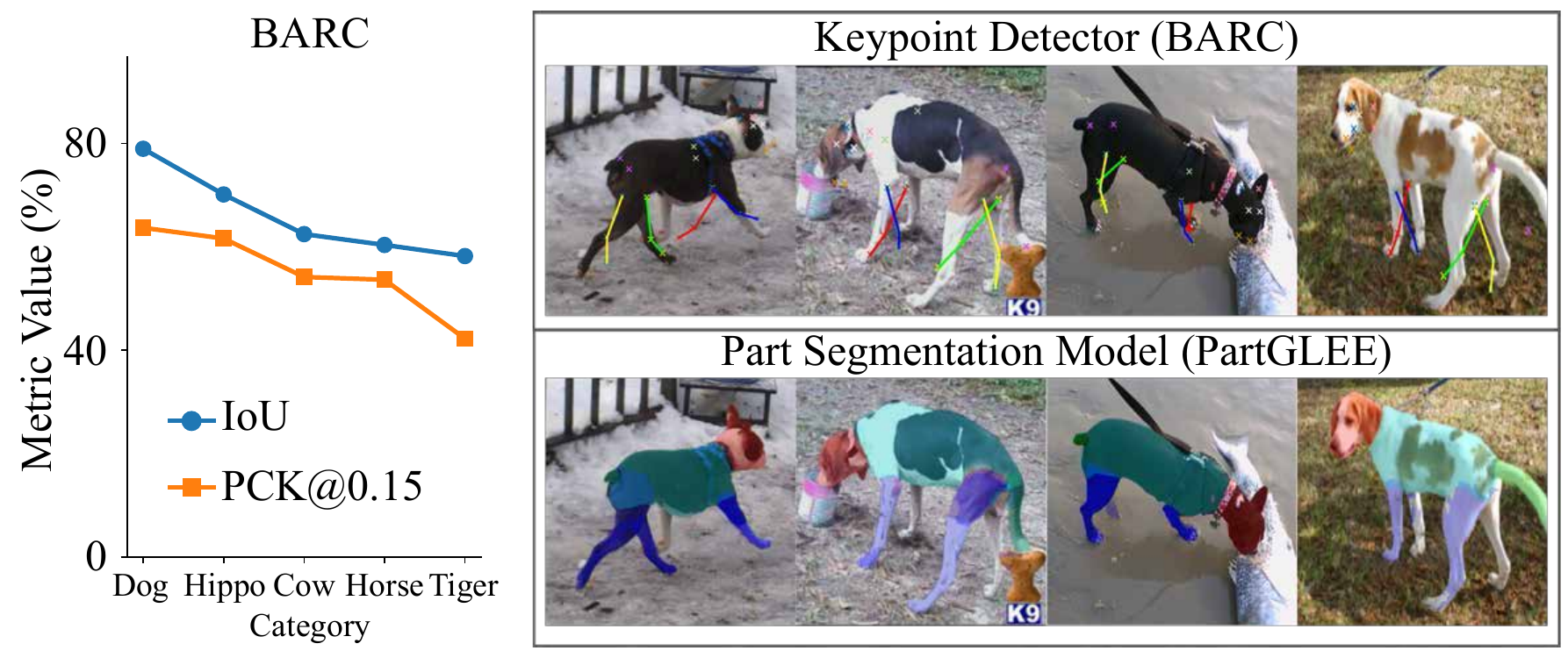}
\vspace{-5pt}
  \caption{\textbf{Left}: Evaluation of the keypoint detector from BARC~\cite{rueegg2022barc} on the Animal3D dataset~\cite{xu2023animal3d}, using IoU and the 2D Percentage of Correct Keypoints (PCK) metric. The category is the super-category defined in Animal3D. \textbf{Right}: Visualization of keypoint detector BARC and part segmentation model PartGLEE~\cite{li2024partglee}. The leg keypoints are distinguished by connecting them with lines. }
  \vspace{-10pt}
\label{fig:detector}
\end{figure}

Recent differentiable rendering methods~\cite{ravi2020accelerating, liu2019soft} have shown promise in obtaining high-quality animatable 3D animals from monocular videos~\cite{sabathier2024animal,yang2023reconstructing}. Among them, model-based methods~\cite{ruegg2023bite,rueegg2022barc} leveraging predefined category template meshes, e.g., SMAL~\cite{zuffi20173d}, the quadruped equivalent of SMPL~\cite{loper2023smpl}, achieve impressive performance. However, these methods require sparse semantic keypoint annotations as shown in Fig.~\ref{fig:introduction} (a) to fit the template meshes to images. 
Acquiring such keypoints is non-trivial. In practice, they are often \textbf{manually annotated}, which significantly increases the cost and restricts scalability to in-the-wild or casual videos.
A common workaround is to employ a keypoint detector~\cite{rueegg2022barc,ruegg2023bite,sabathier2024animal}, but training a reliable detector requires a large dataset with accurately labeled keypoints. Unfortunately, such annotated data is scarce for animals~\cite{xu2023animal3d}, making it \textbf{challenging to train robust detectors}. To assess the effectiveness of this strategy, we evaluate a widely used keypoint detector from BARC~\cite{rueegg2022barc} using manual annotations from the quadruped dataset Animal3D~\cite{xu2023animal3d}. 
We report standard metrics, including IoU and 2D Percentage of Correct Keypoints (PCK), following the thresholds used in BARC. As shown in Fig~\ref{fig:detector} (Left), the detector shows inconsistent accuracy across different animal categories. Besides, Fig.\ref{fig:detector} (Right) shows that it often produces biased predictions, especially for the legs, and lacks the robustness seen in well-trained models such as PartGLEE~\cite{li2024partglee}.
These issues raise a critical question about the sustainability and scalability of current animal reconstruction pipelines:
\begin{mdframed}[backgroundcolor=gray!8]
\begin{minipage}{\linewidth}
\vspace{-0.5pt}
Can we achieve 4D reconstruction from videos \textbf{without} manually annotating sparse keypoints?
\vspace{-0.5pt}
\end{minipage}
\end{mdframed}

To address this limitation, we propose 4D-Animal, a novel approach that eliminates the need for sparse semantic keypoint correspondences. Our key insight is that such sparse keypoints are not necessary, as general image features can provide the necessary guidance for SMAL fitting. 

4D-Animal is designed as an optimization-based method that integrates coarse-to-fine alignment strategies to enhance 4D reconstruction accuracy. It employs a dense feature network to project 2D representation to SMAL parameters across a video. This dense representation not only enhances robustness but also accelerates convergence. Besides, unlike traditional methods that rely on manually annotated keypoints for SMAL fitting, 4D-Animal integrates hierarchical alignment cues that can be obtained by well-trained 2D models~\cite{ravi2024sam,li2024partglee,Neverova2020ContinuousSurfaceEmbeddings,doersch2024bootstap}. We apply object masks for global silhouette consistency, semantic part masks as shown in Fig.~\ref{fig:introduction} (a) for coarse region correspondence, pixel-to-vertex correspondence for fine-grained alignment, and tracking information as shown in Fig.~\ref{fig:introduction} (b) to ensure motion coherence. By incorporating these hierarchical alignment cues, 4D-Animal ensures accurate per-frame pose estimation while maintaining coherent motion across the video. 

\begin{figure}[t]
  \centering
\includegraphics[width=0.89\linewidth]{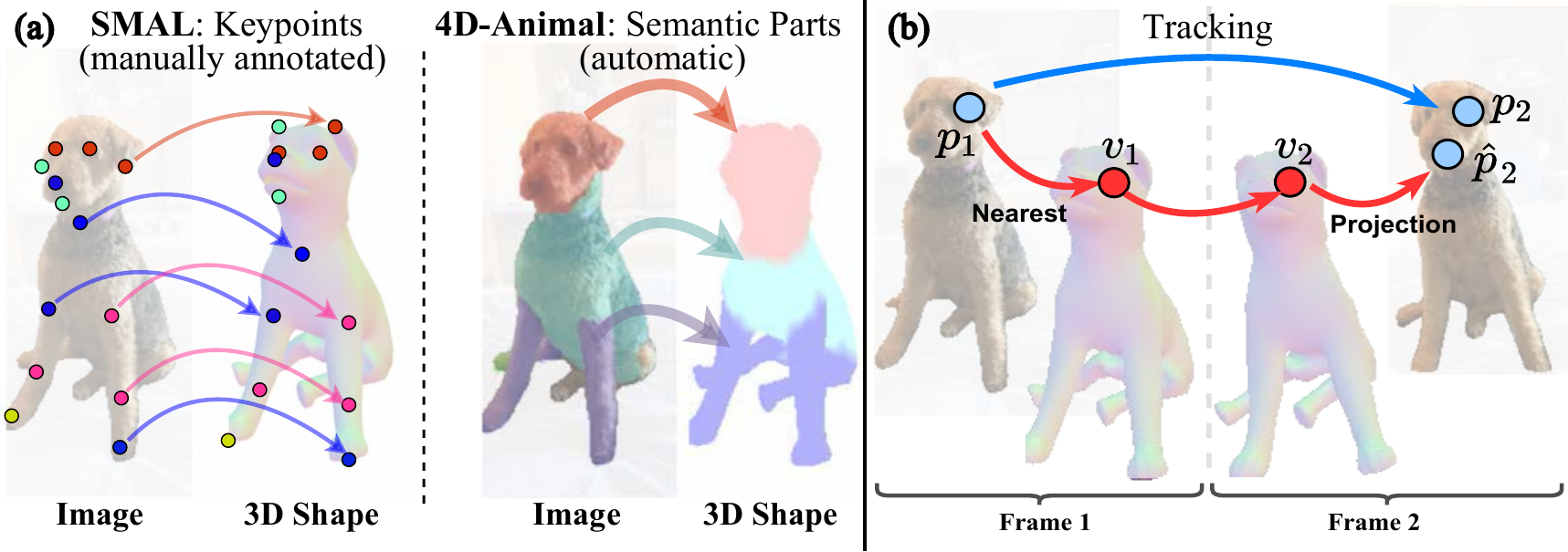}
  \caption{(a) \textbf{Left}: Keypoint alignment (manually annotated) . \textbf{Right}: Part-level alignment (automatic) using semantic masks divided into head, body, feet, and tail, which are mapped to corresponding SMAL mesh regions. (b) Temporal tracking alignment. A tracked 2D point $p_1$ in Frame 1 is matched to $p_2$ in Frame 2. The nearest vertex $v_1$ to $p_1$ is found, and its corresponding vertex $v_2$ is located via SMAL topology. The loss is computed between $p_2$ and the projection of $v_2$. 
  }
  \vspace{-15pt}
\label{fig:introduction}
\end{figure}

We implement 4D-Animal on the widely used dog-specific SMAL model~\cite{ruegg2023bite} to ensure a deep analysis. Comprehensive experiments show that without manual annotations, 4D-Animal achieves state-of-the-art performance on benchmark datasets while improving fitting efficiency, demonstrating the potential to scale up 3D animatable animal reconstruction. 
In summary, our contributions are shown as follows:
\begin{itemize}
     \item We design a dense feature network that maps 2D image representations to SMAL parameters, improving the efficiency and stability of the fitting process.
     \item We propose a hierarchical alignment strategy that combines silhouette, part, pixel-level, and temporal cues to achieve accurate and coherent 3D reconstructions across frames.
     \item Extensive experiments show that 4D-Animal, a simple-yet-efficient framework that removes the need for sparse semantic keypoint annotations, achieves superior performance compared to both model-based and model-free baselines.
\end{itemize}

\section{Related work}
\label{sec:formatting}

Reconstructing animatable 3D objects from videos or single-view images is a long-standing problem in computer vision. While recent methods have obtained impressive results on humans~\cite{goel2023humans}, there is comparably less work for animals. Existing approaches for 3D animal reconstruction can be broadly categorized into two main types: model-free and model-based methods~\cite{sabathier2024animal,ruegg2023bite}.

\noindent\textbf{Model-free methods.}
Recent advances in implicit 3D representations and differentiable rendering have driven progress in model-free methods. Without relying on predefined shape templates, these methods require additional supervision, such as large image datasets or video sequences, to infer geometry.
CMR~\cite{cmrKanazawa18} reconstructs 3D birds using keypoints and viewpoints as additional constraints. LASR~\cite{yang2021lasr} and its extensions~\cite{yang2021viser, yang2022banmo, yang2023reconstructing} optimize 3D animal models from monocular videos, leveraging optical flow, masks, and DensePose~\cite{Neverova2020ContinuousSurfaceEmbeddings} for supervision. LASSIE~\cite{yao2022lassie, yao2023hi-lassie} introduces DINO~\cite{dino} features and a generic skeleton to optimize on small-scale image collections. Dove~\cite{wu2023dove} applies an analysis-by-synthesis autoencoder for video-based animal reconstruction, while MagicPony~\cite{wu2023magicpony} and 3D Fauna~\cite{li2024fauna} extend this framework to single images using DINO features. Dynamic Gaussian Mesh~\cite{liu2024dynamic} integrates 3D Gaussian Splatting~\cite{kerbl3Dgaussians} to reconstruct dynamic animal meshes but struggles in monocular settings.
Despite these advancements, model-free methods often fail to accurately capture shape and articulation. They also struggle with extreme poses, occlusions, and limited viewpoint coverage, leading to ambiguities in the reconstructed geometry.

\noindent\textbf{Model-based methods.}
model-based methods effectively handle ambiguities caused by extreme poses, occlusions, and limited viewpoints~\cite{xu2023animal3d}. Their parametric shape and pose representations also facilitate motion and behavior analysis, as demonstrated in human studies~\cite{goel2023humans}. Most model-based methods rely on the SMAL model~\cite{zuffi20173d}, the only parametric 3D articulated shape model for quadrupeds to the best of our knowledge, with a counterpart for birds~\cite{wang21aves}. SMAL has been widely used for reconstructing quadrupeds from single images~\cite{Zuffi:ICCV:2019, biggs2020wldo, li2021coarse, ruegg2023bite, rueegg2022barc}, RGB-D data~\cite{rgbd_dog}, and videos~\cite{biggs2018creatures, sabathier2024animal}. Advanced versions extend its shape space for specific species to improve accuracy~\cite{ruegg2023bite, Zuffi:CVPR:2024}.  
However, model-based methods require 2D keypoint annotations for each frame~\cite{biggs2018creatures}. To reduce this dependency, keypoint detectors trained for specific species, like dogs, are used~\cite{ruegg2023bite,rueegg2022barc,sabathier2024animal}. While they minimize manual effort, they require large labeled datasets for robustness and struggle with rear and side views due to data scarcity~\cite{sabathier2024animal}. Animal Avatars mitigates this with denser features but still relies on sparse keypoints for thin body structures~\cite{sabathier2024animal}.
In contrast, 4D-Animal eliminates the need for manually annotated sparse keypoints. By leveraging hierarchical alignment cues, 4D-Animal enables efficient SMAL fitting from casual videos, making 3D animatable animal reconstruction more scalable and accessible.

\section{Methodology}
\label{sec:method}

Our framework consists of two main stages: (1) mesh reconstruction via SMAL parameter regression, and (2) texture reconstruction using duplex mesh and triplane feature grid. These stages are jointly supervised by hierarchical alignment loss and a perceptual texture loss.

\subsection{Preliminaries}

\noindent\textbf{SMAL model} $M(\beta, \theta)$ parameterizes 3D geometry of meshes using shape parameters $\beta \in \mathbb{R}^{41}$ and pose parameters $\theta \in \mathbb{R}^{105}$ to represent quadruped poses and shapes. Given these parameters, the SMAL model is a differentiable function that outputs a posed 3D mesh $m \in \mathbb{R}^{3889 \times 3}$. Note that the dimensionality of $\beta$, $\theta$, and the vertex number may vary when we use more advanced SMAL models~\cite{ruegg2023bite} for specific species.

\noindent\textbf{SMAL fitting} is the process of fitting the SMAL model to an image by optimizing the shape and pose parameters guided by a combination of 2D keypoints and 2D silhouettes. 
Note that the supervision used in SMAL fitting always includes the terms that encourage parameters to be close to the prior distribution and the constraints to pose parameter $\theta$ with limit bounds.

\begin{figure*}[b]
  \centering
\includegraphics[width=\linewidth]{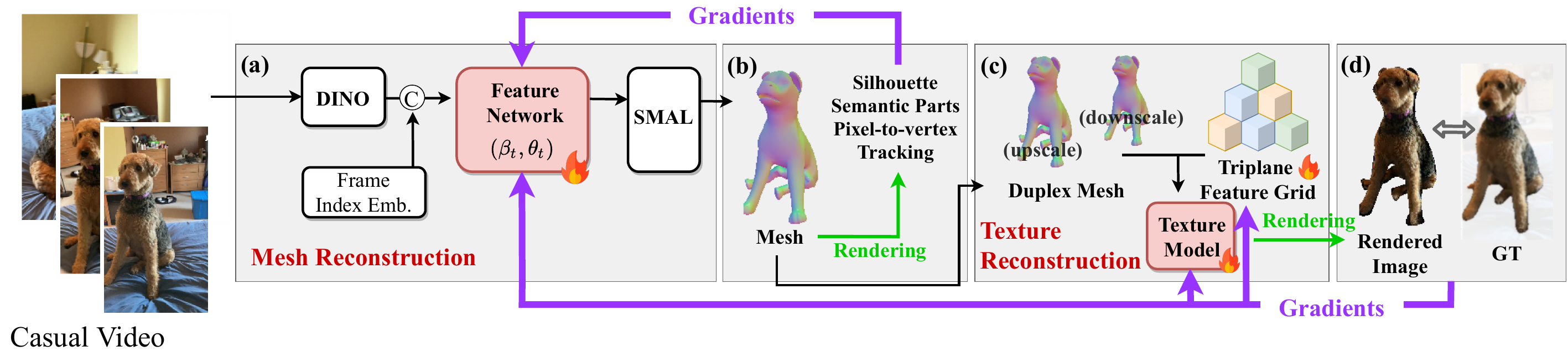}
  \caption{The overview of proposed 4D-Animal. (a) Mesh reconstruction. ``C" stands for concatenation. ``Frame Index Emb." refers to the embedding of the index of a video frame. $(\beta_t, \theta_t)$ are inputs to SMAL for $t$-th frame. The feature network is learnable. (b) Hierarchical geometric alignment. The alignment loss is constructed using silhouette, semantic part, pixel, and tracking information generated by a series of 2D vision pre-trained models. (c) Texture reconstruction. The RGB texture is reconstructed through a learnable texture model and triplane feature grid based on the duplex mesh. (d) Texture loss. }
  \vspace{-15pt}
\label{fig:metohd}
\end{figure*}

\subsection{Model framework}

As shown in Fig.~\ref{fig:metohd}, 4D-Animal reconstructs mesh and texture captured in a monocular video. 
Specifically, given a video $(I_t)^T_{t=1}$ with $T$ frames, we build animatable 3D models which can outputs colored 3D shapes.
For mesh reconstruction, we project image features to SMAL parameters using a feature network. For texture reconstruction, a texture model uses duplex mesh and triplane feature grid to refine textures.

\noindent\textbf{Mesh reconstruction. }We represent the mesh $m_t$ of frame $t$ with SMAL model $M(\beta_t, \theta_t)$.  
For each frame $I_t \in R^{3 \times H \times W}$, we learn a function $f: I_t \mapsto \text{SMAL parameters}$ that maps $I_t$ of a quadruped to a corresponding mesh, capturing the quadruped's shape and pose. As shown in Fig.~\ref{fig:metohd} (a), 
we use a well-trained feature extractor DINO-ViT ~\cite{caron2021emerging} to extract image feature $z_t$. Then a feature network, accepting image feature $z_t$ and positional encoding $e_t$ of frame $I_t$ as input, outputs SMAL parameters. We use shallow multi-layer perceptron as the framework of the feature network, and following D-SMAL~\cite{ruegg2023bite}, limb length scaling factors, translation, and a set of vertices shifts are also applied to SMAL.

\noindent\textbf{Texture reconstruction. }We also aim to reconstruct the texture of mesh $m_t$. However, high-quality texture requires an exact supporting 3D shape, while $m_t$ reconstructed by SMAL has only 3,889 vertices and 7,774 triangular faces, so directly modeling vertex texture lacks expressivity. Therefore, as shown in Fig.~\ref{fig:metohd} (c), we leverage $m_t$ as a scaffold to build duplex mesh and implement the coloring of $m_t$ by using a shallow multi-layer perceptron decoding a learnable triplane feature grid following recent advances in new-view synthesis~\cite{chan2022efficient,cao2024lightplane,sabathier2024animal}. Refer Appendix 1.1 for details.  

\subsection{Learning formulation}
\label{sec:learning}

Most 3D animal reconstruction methods rely on sparse keypoints~\cite{zuffi20173d, biggs2019creatures}, but this assumption breaks down for in-the-wild animals due to anatomical diversity and limited annotations. Instead, we propose a coarse-to-fine, hierarchical supervision strategy that avoids explicit keypoints and leverages weak but dense cues: instance masks, semantic parts, pixel-to-vertex correspondences, and motion tracks. This design enables robust shape, pose, and motion estimation across frames.

\textbf{Hierarchical 4D geometric alignment.}  
As shown in Fig.~\ref{fig:metohd} (b),  we formulate four levels of geometric losses, each progressively refining the fitting. 

- At the \emph{object level}, we use instance masks $u_t$ from SAM~\cite{ravi2024sam} to align the projected mesh silhouette with the image using Chamfer distance:  
  $\mathcal{L}^{\text{obj}} = \sum_{t} D(u_t, P_t v_t)$,  
  where $P_t$ is the projection matrix, and $v_t$ the mesh vertices. This ensures global shape coverage.

- At the \emph{part level}, we exploit semantic part masks $s_t$ from PartGLEE~\cite{li2024partglee}, segmented into head, body, feet, and tail. We sample $N_s$ points $p_t^s$ (proportional to part area) and associate them with annotated SMAL mesh regions as $v_t^s$, forming the loss:  
  $\mathcal{L}^{\text{part}} = \sum_t \|p_t^s - P_t v_t^s\|^2$.  
  This introduces coarse semantic structure beyond silhouettes. 

- At the \emph{pixel level}, we use CSE~\cite{Neverova2020ContinuousSurfaceEmbeddings} to obtain dense pixel-to-vertex correspondences. Since CSE and SMAL use different meshes, we align their coordinate systems via Zoom-Out~\cite{melzi2019zoomout}, and supervise foreground pixels $p_t^c$ against projected mesh points $v_t^c$:  
  $\mathcal{L}^{\text{pix}} = \sum_t \|p_t^c - P_t v_t^c\|^2$.  
  This allows fine-grained alignment, particularly for details like limb orientation.

- At the \emph{temporal level}, we use 2D trajectories from BootsTAP~\cite{doersch2024bootstap}. Given a track $p_t \rightarrow p_t'$ between frames $t$ and $t'$, we project the corresponding mesh vertex $v_t \rightarrow v_t'$ and minimize:  
  $\mathcal{L}^{\text{time}} = \sum_t \|p_t' - P_t' v_t'\|^2$.  
  This enforces motion coherence across frames. 

Each geometric loss targets a specific level of alignment, and their combination is crucial for accurate 4D reconstruction. While pixel-to-vertex correspondences provide fine-grained cues, they are often noisy in practice, leading to silhouette and motion errors. Object- and part-level losses add silhouette guidance, and temporal alignment ensures cross-frame consistency. Together, these complementary losses form a robust framework. Details of each alignment level are provided in Appendix 1.3.

\textbf{Texture loss and full objective.}  
To encourage visual realism, we add a perceptual texture loss using LPIPS~\cite{zhang2018unreasonable} between the masked image $I_t^u$ and the textured mesh rendering $T(m_t)$:  
$\mathcal{L}^{\text{tex}} = \sum_t \text{LPIPS}(I_t^u, T(m_t))$.

In summary, the training objective $\mathcal{L}$ is thus
\begin{equation}
  \mathcal{L} = \mathcal{L}^{\text{geo}} + \lambda^{\text{tex}} \mathcal{L}^{\text{tex}} + \mathcal{R},
\end{equation}

\noindent where $\lambda^{\cdot}$ are hyper-parameters that control the importance of different loss terms and  $\mathcal{L}^{\text{geo}} =  \sum_{i \in \{\text{obj}, \text{part}, \text{pix}, \text{time}\}}\lambda^i \mathcal{L}^i$ summarizes the four geometric losses. In addition to the geometry and texture loss functions $\mathcal{L}^{\text{geo}}, \mathcal{L}^{\text{tex}}$, we incorporate shape regularization~\cite{igarashi2005rigid,sorkine2004laplacian}, volume consistency, and laplacian smoothing~\cite{nealen2006laplacian,desbrun1999implicit} during model fitting to enhance stability and ensure a coherent reconstruction and summarize such regularizers as $\mathcal{R}$.

\subsection{Camera initialization}
\label{sec:camera}

For a non-rigid reconstruction for each frame, a bad initialization of the camera parameters inevitably leads to failure. Although many existing works~\cite{wang2024vggsfm,smith2024flowmap} can reconstruct the camera poses from a set of unconstrained 2D images, such camera initialization is rough, adding difficulties to our task.

To this end, we leverage pixel-to-vertex correspondences to refine the camera parameter before model fitting. Specifically, for $t$-th frame $I_t$, we use part-level points $p_t^s$ and pixel-level points $p_t^c$, and their corresponding vertices $v_t^s$ and $v_t^c$. Then using an efficient PnP algorithm EPnP-RANSAC~\cite{lepetit2009ep} to find a camera position that minimizes re-projection error between the given uncalibrated 2D points $p_t^s, p_t^c$ and the corresponding 3D points $v_t^s, v_t^c$. 
We find that both part-level and pixel-level correspondences are essential for camera initialization. While pixel-level correspondence provides fine-grained alignment, it often introduces some bias and fails in some cases. In contrast, part-level correspondence, though coarse, offers reliably accurate part information. These two types of correspondences complement each other, balancing precision and robustness.

\section{Experiments}

\subsection{Experimental settings}

\noindent\textbf{Datasets. }We evaluate all models on COP3D~\cite{sinha2023common}, a public dataset featuring fly-around videos of pets, with annotated camera parameters and object masks. Following~\cite{sabathier2024animal}, we select a subset of 50 dog videos capturing diverse poses, motions, and textures. Each test video contains 200 frames, which we split into training and test sets by considering contiguous blocks of 15 frames as train, interleaved by blocks of 5 frames as test. All models are evaluated at a resolution of 256$\times$256. We also use casual videos with depth maps of dogs from TracksTo4D~\cite{kasten2024fast} to evaluate the reconstructed quality.

\noindent\textbf{Metrics. }We use three metrics to evaluate shape and texture quality: silhouette accuracy via Intersection-over-Union (IoU), and appearance fidelity via Peak Signal-to-Noise Ratio (PSNR) and Learned Perceptual Image Patch Similarity (LPIPS)~\cite{zhang2018unreasonable}. We also report the worst 5\% variant IoUw5 and PSNRw5. Besides, we use two depth metrics Absolute Relative Error (Abs Rel) and Delta ($\delta$) accuracy to evaluate the reconstructed structure.

\noindent\textbf{Baselines. } We compare 4D-Animal against the state-of-the-art model-based counterparts, including Avatars~\cite{sabathier2024animal}, BARC~\cite{rueegg2022barc}, and BITE~\cite{ruegg2023bite}, as well as the model-free approach RAC~\cite{yang2023reconstructing}. The results of BARC, BITE, and RAC are from Avatars~\cite{sabathier2024animal}. Since only Avatars and RAC incorporate texture reconstruction, texture quality comparisons are limited to these two methods, whereas BITE and BARC are evaluated solely on 3D shape accuracy.

\begin{table}[b]
  \caption{Average results on 50 sequences from COP3D. Quality of pose estimates is measured by IoU, IoUw5; texture quality is measured by PSNR, PSNRw5, LPIPS. Note that BARC and BITE only estimate pose and hence we cannot evaluate texture quality, indicated by `-'. *: we re-run the official codebase and collect the results. SK: training with sparse keypoints. Bold and underline indicate the best results and the second-best results, respectively. We also report the percentage improvement of 4D-Animal over SOTA method.}
  \centering
  \renewcommand{\arraystretch}{1.3}
  \resizebox*{0.99\linewidth}{!}{
    \begin{tabular}{ll|ccccc}
    \toprule
    Models & SK & IoU ↑  & IoUw5 ↑ & PSNR ↑ & PSNRw5 ↑ & LPIPS ↓ \\
    \midrule
    BARC~\cite{rueegg2022barc} & \checkmark & 0.75  & 0.47  & -     & -     & - \\
    BITE~\cite{ruegg2023bite} & \checkmark & 0.81  & 0.59  & -     & -     & - \\
    RAC~\cite{yang2023reconstructing} &  & 0.76  & 0.52  & \textbf{21.86} & \underline{17.51} & 0.164 \\
    Avatars*~\cite{sabathier2024animal} & \checkmark & \underline{0.81}  & \underline{0.66}  & 19.91 & 17.25 & \underline{0.073} \\
    \midrule
    4D-Animal & & \textbf{0.84} ({\color{red}{$\uparrow$ 3.70\%}}) & \textbf{0.71} ({\color{red}{$\uparrow$ 7.58\%}}) & \underline{21.28} ({\color{red}{$\uparrow$ 7.38\%}}) & \textbf{18.17} ({\color{red}{$\uparrow$ 5.33\%}}) & \textbf{0.061} ({\color{red}{$\uparrow$ 16.44\%}}) \\
    \bottomrule
    \end{tabular}%
  }
  \label{tab:main}%
\vspace{-10pt}
\end{table}%

\begin{figure*}[t]
  \centering
\includegraphics[width=0.79\linewidth]{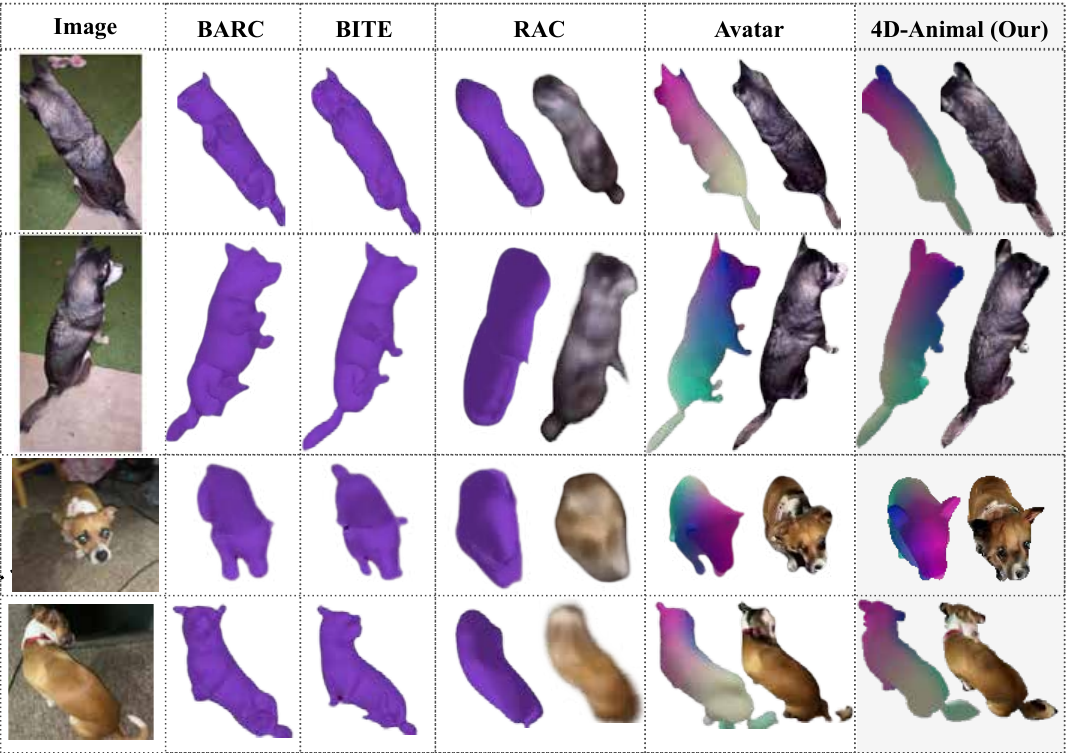}
  \caption{Qualitative comparison. We compare our 4D-Animal to baselines by selecting images from different viewpoints in videos.}
\vspace{-10pt}
\label{fig:visualization}
\end{figure*}

\subsection{Overall performance}

Table~\ref{tab:main} shows that 4D-Animal outperforms other methods on most metrics, even without using sparse keypoints. Qualitative comparisons in Fig.~\ref{fig:visualization} further support this. BARC and BITE struggle with temporal consistency, leading to poor IoUw5 scores. RAC, a model-free method, produces the worst geometry, highlighting the importance of model-based fitting. Avatars performs well but relies on sparse keypoints. It often fails on non-frontal views since keypoints provide weak supervision for rear and side perspectives~\cite{sabathier2024animal}. In contrast, 4D-Animal, without using keypoints, achieves the best overall reconstruction. 4D-Animal improves both efficiency and scalability by removing the need for unreliable keypoint annotations. Additional comparisons on reconstruction quality and temporal consistency are provided in Appendix 2, along with reconstruction results for other animal types.

\begin{table}[t]
  \caption{Quantitative evaluation of reconstructed geometry using depth-based metrics. We report Absolute Relative Error (Abs Rel) and threshold accuracy metrics ($\delta < 1.25^n, n \in \{1, 2, 3\}$). }
  \centering
  \resizebox*{0.60\linewidth}{!}{
    \begin{tabular}{lcccc}
    \toprule
    Models & Abs Rel ↓ & $\delta < 1.25$ ↑ & $\delta < 1.25^2$ ↑ & $\delta < 1.25^3$ ↑ \\
    \midrule
    Avatars~\cite{sabathier2024animal} & 0.427 & 0.651 & 0.869 & 0.907 \\
    \rowcolor[rgb]{ .851,  .851,  .851} 4D-Animal & 0.279 & 0.739 & 0.915 & 0.951 \\
    \bottomrule
    \end{tabular}%
  }
  \label{tab:depth}%
  \vspace{-10pt}
\end{table}%

\begin{figure}[t]
  \centering
\includegraphics[width=0.60\linewidth]{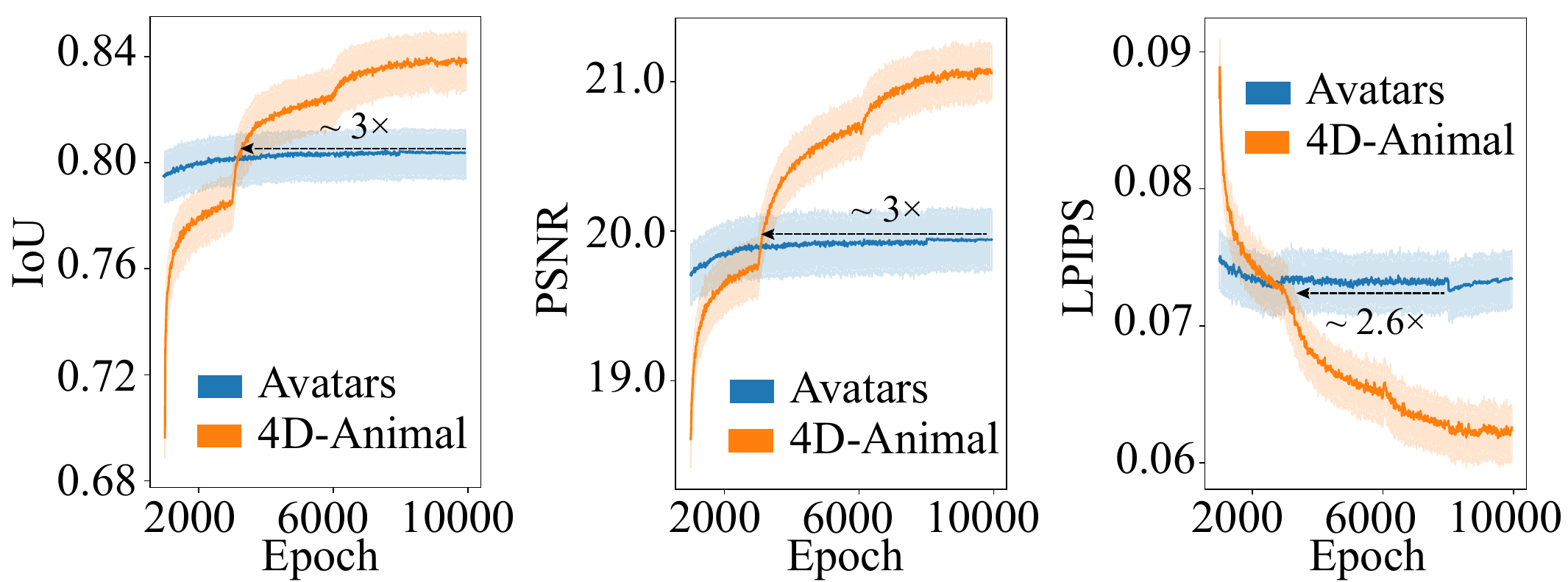}
\vspace{-5pt}
  \caption{Comparison of training efficiency between 4D-Animal and Avatars. }
  \vspace{-10pt}
\label{fig:efficiency}
\end{figure}

\begin{figure}[t]
  \centering
\includegraphics[width=0.70\linewidth]{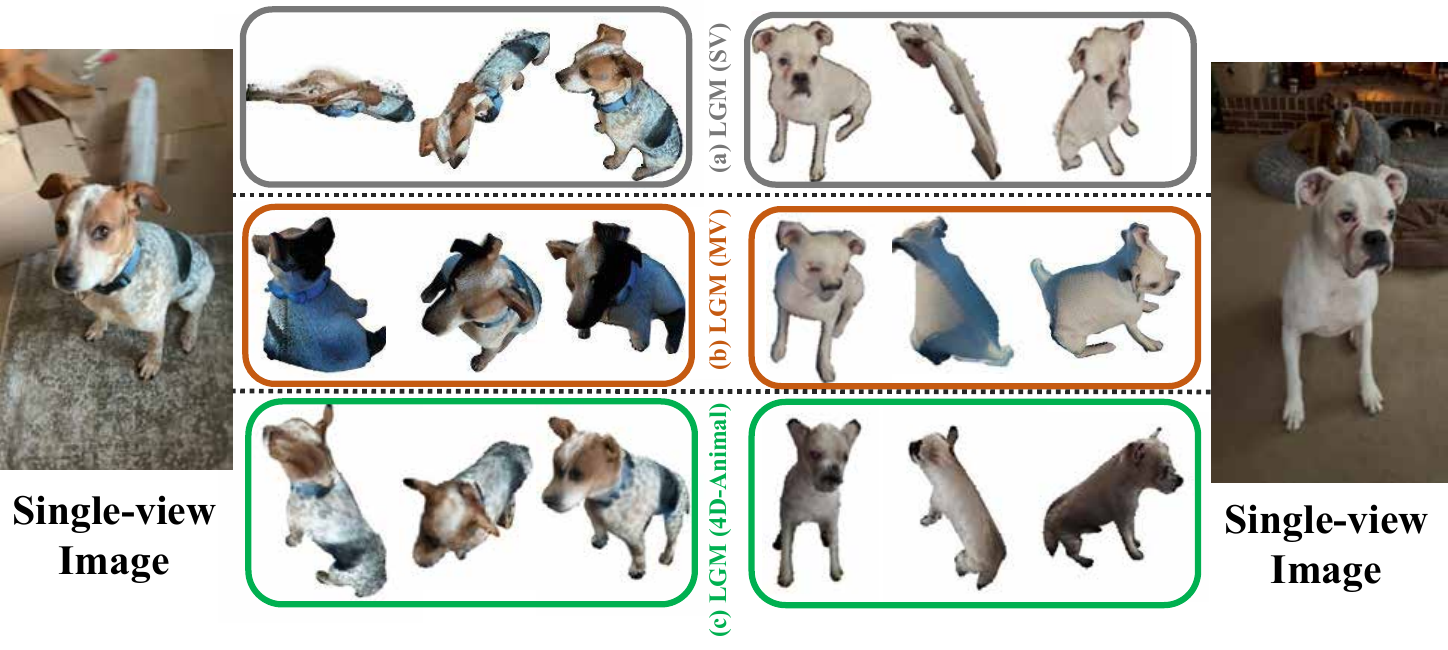}
\vspace{-8pt}
  \caption{Comparison of the single-view image to 3D generation based on LGM~\cite{tang2024lgm} (SV: single-view only, MV: multi-view using MVDream~\cite{shi2023mvdream}). Efficient finetuning with 4D-Animal greatly reduces the sim-to-real gap.}
  \vspace{-12pt}
\label{fig:image-to-3D}
\end{figure}

\begin{figure}[b]
  \centering
\includegraphics[width=0.75\linewidth]{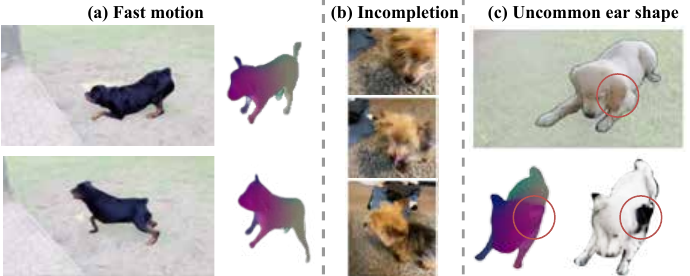}
\vspace{-5pt}
  \caption{Failure cases by 4D-Animal. }
  \vspace{-10pt}
\label{fig:failure}
\end{figure}

\section{Analysis}
\label{sec:analysis}

In this section, we analyze several key research questions (RQs). First, we quantitatively evaluate 4D-Animal itself, focusing on reconstruction quality and training efficiency. We then explore the feasibility of leveraging synthetic data from 4D-Animal to assist other 3D task. Finally, we provide a summary and explanations for 4D-Animal's failures.

\noindent\textbf{RQ1. Quantitative evaluation of geometry structure reconstructed by 4D-Animal.}

\noindent To assess the geometric accuracy of 4D-Animal, we use the dog videos curated by TracksTo4D~\cite{kasten2024fast}. These videos, captured with an RGB-Depth sensor, provide depth maps as ground truth for evaluating reconstruction quality. For each method, we reconstruct meshes per frame using each method, then we can calculate a synthetic depth map and follow previous work \cite{luo2020consistent} to account for the unknown global scale factor. Specifically, we compute the ratio of synthetic depth to ground-truth depth at each frame and use the median value as the scale factor for alignment, ensuring a fair comparison. As shown in Table~\ref{tab:depth}, 4D-Animal outperforms the state-of-the-art baseline, Avatars, demonstrating its superior ability to reconstruct accurate 3D geometry.

\begin{table}[t]
  \caption{Ablation study on the impact of hierarchical alignment.}
  \centering
  \resizebox*{0.69\linewidth}{!}{
    \begin{tabular}{lccccc}
    \toprule
    w/o   & IoU ↑  & IoUw5 ↑ & PSNR ↑ & PSNRw5 ↑ & LPIPS ↓ \\
    \midrule
    $\mathcal{L}^{\text{obj}}$  & 0.51  & 0.38  & 17.42 & 14.84 & 0.115 \\
    $\mathcal{L}^{\text{part}}$ & 0.84  & 0.70  & 21.17 & 17.90 & 0.063 \\
    $\mathcal{L}^{\text{pix}}$ & 0.83  & 0.68  & 20.85 & 18.01 & 0.063 \\
    $\mathcal{L}^{\text{time}}$ & 0.84  & 0.69  & 21.06 & 18.28 & 0.062 \\
    $\mathcal{L}^{\text{tex}}$  & 0.84  & 0.70  & 12.94 & 10.31 & 0.147 \\
    \midrule
    4D-Animal & 0.84  & 0.71  & 21.28 & 18.17 & 0.061 \\
    \bottomrule
    \end{tabular}%
  }
  \label{tab:alignment}%
  \vspace{-10pt}
\end{table}%

\begin{table}[htbp]
  \centering
  \caption{Evaluation of camera initialization strategies introduced in Sec.~\ref{sec:camera}. EPnP~\cite{lepetit2009ep} is an PnP algorithm to find a camera position, and RANSAC enhances robustness by filtering outliers.}
  \resizebox*{0.69\linewidth}{!}{
    \begin{tabular}{l|c|ccc|ccc}
    \toprule
    Initialization & Org   & \multicolumn{3}{c|}{EPnP} & \multicolumn{3}{c}{EPnP-RANSAC} \\
    \midrule
    Part  &       & \checkmark     &       & \checkmark     & \checkmark     &       & \cellcolor[rgb]{.851, .851, .851} \checkmark \\
    Pixel &       &       & \checkmark     & \checkmark    &       & \checkmark     & \cellcolor[rgb]{.851, .851, .851} \checkmark \\
    \midrule
    IoU ↑ & 0.135 & 0.306 & 0.399 & 0.400 & 0.148 & 0.451 & \cellcolor[rgb]{.851, .851, .851}0.464 \\
    IoUw5 ↑ & 0.077 & 0.199 & 0.286 & 0.295 & 0.102 & 0.331 & \cellcolor[rgb]{.851, .851, .851} 0.342 \\
    \bottomrule
    \end{tabular}%
    }
  \label{tab:init}%
\end{table}%

\noindent\textbf{RQ2. Training efficiency of 4D-Animal.}

\noindent As shown in Fig.~\ref{fig:efficiency}, we use 50 videos from COP3D to compare the training efficiency of Avatars and 4D-Animal across three key metrics. Our method achieves target metric values faster than Avatars, with IoU speedup of up to 3 $\times$. This highlights the efficiency of our feature network design and learning formulation, demonstrating 4D-Animal's ability to accelerate training while maintaining high reconstruction quality. Refer Appendix 2.1 for more details about our training procedure.

\noindent\textbf{RQ4.  Can the 3D assets generated by 4D-Animal assist in other 3D tasks?}

\noindent To explore this question, we assess 3D assets from 4D-Animal on the popular task, image-to-3D generation. We observe that current 3D generative models trained on synthetic datasets often fail when processing real-world animal images. Taking LGM~\cite{tang2024lgm} as a representative example, which is trained on the large-scale Objaverse dataset~\cite{deitke2023objaverse}, the generated results suffer huge performance loss due to the domain gap. As shown in Fig.~\ref{fig:image-to-3D} (a), the model cannot predict a reasonable depth and degrades to a flat structure when using the single-view image as input. Besides, as shown in Fig.~\ref{fig:image-to-3D} (b), leveraging multi-view images generated from existing 2D diffusion models like MVDream~\cite{shi2023mvdream} is still hard to maintain the 3D consistency while conforming faithfully to the original image. In contrast, when fine-tuning LGM with 3D assets reconstructed by our 4D-Animal, it is possible to both utilize the 3D prior learned on large synthetic datasets and effectively narrow the gap of sim-to-real. Fig.~\ref{fig:image-to-3D} (c) demonstrates the high quality of our generated results from the single-view \textit{unseen} image, and 4D-Animal can benefit diverse generative models with different representations.

\noindent\textbf{RQ4.  Failure cases by 4D-Animal.}

\noindent 4D-Animal encounters challenges primarily in scenarios with fast motion, incomplete animal visibility, and uncommon shape of the ears, as shown in Fig.~\ref{fig:failure}. These factors make it difficult to use well-trained models to accurately capture hierarchical alignment cues.
As shown in Fig.~\ref{fig:failure} (a), fast motion leads to motion blur, which limits the clarity of image features. While 4D-Animal can still reconstruct motion-consistent shapes, the fine-grained shape details may not perfectly align with the actual structure, resulting in some misalignment. Additionally, as shown in Fig.~\ref{fig:failure} (b), when an animal is only partially visible in the video, 4D-Animal struggles to infer and fit the occluded regions, sometimes resulting in inaccurate reconstruction. Besides, as shown in Fig.~\ref{fig:failure} (c), 4D-Animal struggles with handling uncommon ear shapes, particularly when the ears overlap with the body, due to the lack of fine-grained supervision.
These limitations stem from inherent challenges in the input data, highlighting the need for future improvements in handling motion blur and occlusions.

\section{Ablation study}

\noindent\textbf{The efficiency of hierarchical alignment.}
To evaluate 4D-Animal’s learning formulation, we perform an ablation study by individually removing different levels of alignment introduced in Sec.~\ref{sec:learning}. The results in Table~\ref{tab:alignment} reveal that object-level alignment ($\mathcal{L}^{\text{obj}}$) is crucial for IoU performance, while texture alignment ($\mathcal{L}^{\text{tex}}$) strongly influences texture-related metrics. Removing part-level ($\mathcal{L}^{\text{part}}$), pixel-level ($\mathcal{L}^{\text{pix}}$), or temporal ($\mathcal{L}^{\text{time}}$) alignment leads to a slight performance drop, indicating that each component contributes positively to the overall reconstruction quality. These findings confirm the effectiveness of our hierarchical alignment strategy. Refer Appendix 2 for the qualitative comparison of this ablation study.

\noindent\textbf{The effect of camera initialization.}
To evaluate the effectiveness of our camera initialization strategy introduced in Sec.~\ref{sec:camera}, we compare EPnP and EPnP-RANSAC~\cite{lepetit2009ep}, as well as the impact of incorporating part- and pixel-level correspondences. The results, presented in Table~\ref{tab:init}, show that EPnP-RANSAC is more efficient than EPnP, and the combination of both part- and pixel-level correspondences yields the best performance.

\section{Limitations}
4D-Animal  effectively eliminates the unreliable sparse semantic keypoints commonly used in previous methods, but similar to existing approaches, 4D-Animal still cannot achieve perfect reconstruction for any casual video, especially in cases involving rapid animal motion, incomplete visibility, or uncommon body shapes. These challenges are largely inherent in in-the-wild data and remain open problems in the field. Improving robustness under such conditions is an important direction for future work. In addition, our method integrates several pretrained models, which introduces a modest increase in computational overhead during training. Nonetheless, this cost is negligible compared to the overall training expense dominated by backpropagation.

\section{Conclusion}
In this work, we introduce 4D-Animal, an efficient framework for reconstructing animatable 3D animals from monocular videos without keypoint correspondence. By leveraging well-trained 2D visual features and hierarchical alignment cues, 4D-Animal eliminates unreliable keypoint annotations while achieving robust pose estimation and shape optimization. Our method not only outperforms existing model-based approaches in accuracy and motion consistency but also demonstrates scalability for in-the-wild video applications. With the removal of dependency on manually annotated keypoints, 4D-Animal has the potential to be a practical solution for large-scale 3D animal modeling, opening avenues for adoption in animation, biology, and VR.

{
\small
\bibliographystyle{unsrt}
\bibliography{main}
}

\appendix

\newpage

\section{Additional details of 4D-Animal}

\subsection{The details of texture model}
High-quality texture requires an exact supporting 3D shape. However, mesh $m_t$ reconstructed by SMAL has only 3,889 vertices and 7,774 triangular faces, so directly modeling vertex texture lacks expressivity. Therefore, following recent advances in new-view synthesis of humans~\cite{chan2022efficient,cao2024lightplane} and the design of animals~\cite{sabathier2024animal}, we leverage $m_t$ as a scaffold supporting a more accurate implicit radiance field.

We first extend SMAL mesh into a 3D volume by interpolating arbitrary face attributes between a canonical SMAL mesh and its deformed instances at multiple scales. Specifically, as shown in Fig.~\ref{fig:metohd} (c), given a posed SMAL mesh $m_t$ parameterized by shape and pose, we construct both an upscaled and a downscaled version, ensuring a local geometric neighborhood around the canonical surface. By rasterizing both meshes under a differentiable renderer, we identify per-pixel surface intersections and interpolate their precise 3D locations using precomputed canonical vertex structures. This results in a structured mapping, where each pixel is associated with two 3D points, providing a robust foundation for texture reconstruction. 

Then, the rendering process implemented by Lightplane~\cite{cao2024lightplane} follows a volumetric Emission-Absorption model inspired by NeRF, integrating a hybrid representation to efficiently reconstruct appearance. Given a set of rays, each sampled at multiple points along its path, the model estimates per-point features by querying a learned representation (triplane feature grid as shown in Fig.~\ref{fig:metohd} (c)) and refines them via MLP-based decoding. The transmittance along each ray is computed to model light attenuation, ensuring physically consistent color composition at the image plane. By leveraging a structured decomposition of feature extraction and transformation, the approach maintains compatibility with powerful hybrid representations while significantly reducing memory overhead. This enables high-fidelity color reconstruction with improved computational efficiency.

\begin{figure*}[h]
  \centering
\includegraphics[width=\linewidth]{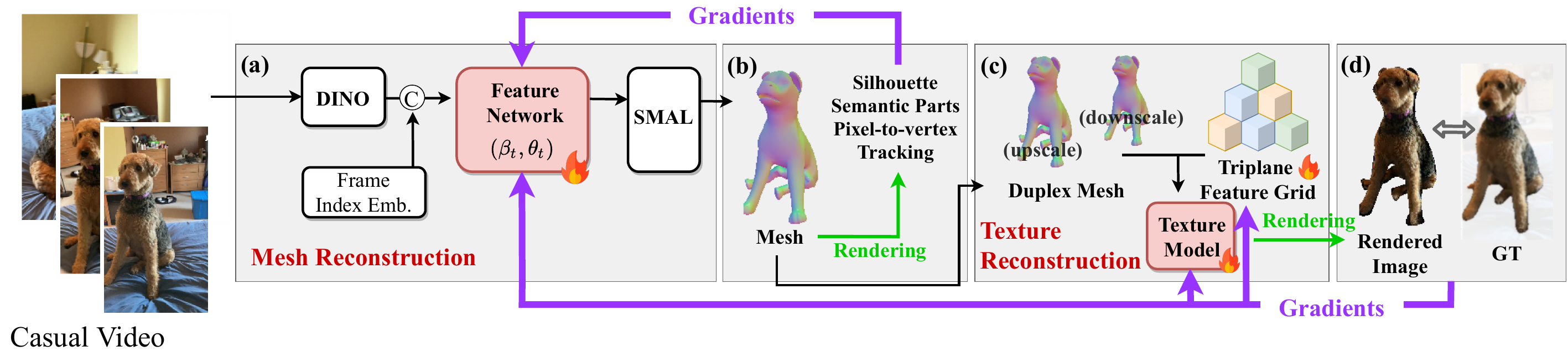}
  \caption{The overview of proposed 4D-Animal. (a) Mesh reconstruction. ``C" stands for concatenation. ``Frame Index Emb." refers to the embedding of the index of a video frame. $(\beta_t, \theta_t)$ are inputs to SMAL for $t$-th frame. The feature network is learnable. (b) Hierarchical geometric alignment. The alignment loss is constructed using silhouette, semantic part, pixel, and tracking information generated by a series of 2D vision pre-trained models. (c) Texture reconstruction. The RGB texture is reconstructed through a learnable texture model and triplane feature grid based on the duplex mesh. (d) Texture loss. }
  \vspace{-15pt}
\label{fig:metohd}
\end{figure*}

\subsection{The details of feature network}

The architecture of the feature network shown in Fig.~\ref{fig:metohd} (a) is highly streamlined. We use a single-layer linear network as the reduction layer to project the 2D representation and 8-dimensional positional encoding into 64-dimensional vectors. This is followed by a single-layer linear network with the same dimensionality, serving as the intermediate feature layer. Finally, another single-layer linear network acts as the output layer, mapping the 2D representation to the SMAL parameters.

\subsection{The details of learning formulation}

\subsubsection{Object-level alignment }

For in-the-wild videos, object masks can be easily generated by a well-trained segmentation model, i.e., SAM~\cite{ravi2024sam}. These masks provide coarse supervision by encouraging the projected mesh silhouette to match the animal’s outline in the image. Specifically, we use instance masks $u_t$ to enforce alignment between the projected mesh and the animal silhouette via the Chamfer distance: $\mathcal{L}^{\text{obj}} = \sum_{t} D(u_t, P_t v_t)$, where $P_t$ is the projection matrix, and $v_t$ the mesh vertices. This ensures global shape coverage.

However, instance masks ignore the semantic consistency, for example, neglecting the difference between the head and tail of the dog as shown in Fig.~\ref{fig:training} (b). This ambiguity can lead to incorrect pose estimation and unrealistic deformations. To address this limitation, we introduce additional coarse-to-fine guidance to enhance semantic alignment.

\subsubsection{Points sampling for part-level alignment}  
At the \emph{part level}, we utilize semantic part masks $s_t$ predicted by PartGLEE~\cite{li2024partglee}, which segment the animals into head, body, feet, and tail as shown in Fig.~\ref{fig:introduction} (a). In the COP3D videos, most frames contain clear head and body regions, but the feet and tail may be missing or occluded. To reduce the impact of inaccurate predictions, we retain feet and tail masks only when their confidence scores from PartGLEE exceed 0.3. Additionally, for blurry frames where no parts are detected, we reuse the part masks from the previous frame.

For each semantic mask $s_t$, we first compute the area of each part (head, body, feet, tail), and allocate the number of sampled points $N_s$ proportionally. For example, the body may receive more samples than the head if it occupies a larger area. We then sample 2D points $p_t^s$ uniformly within each part mask by randomly selecting pixel locations with equal probability. Each sampled point is assigned a semantic region label, and we identify the corresponding SMAL mesh vertices $v_t^s$ using the part annotations defined on the SMAL model. Formally, the part-level loss is formulated as:  
  $\mathcal{L}^{\text{part}} = \sum_t \|p_t^s - P_t v_t^s\|^2$.  In our experiments, $N_s$ is set as 200.

\subsubsection{Pixel-to-vertex alignment via CSE to SMAL}  
At the \emph{pixel level}, we use CSE~\cite{Neverova2020ContinuousSurfaceEmbeddings} to obtain dense pixel-to-vertex correspondences. CSE produces a per-pixel coordinate map aligned to a predefined 3D template. Since this template differs from SMAL, we employ Zoom-Out~\cite{melzi2019zoomout} to construct a functional map between the two surfaces. Once the mapping is obtained, we transfer CSE outputs into the SMAL vertex space. Formally, we supervise the dense foreground pixel coordinates $p_t^c$ by enforcing consistency with the corresponding projected mesh vertices $v_t^c$ via the following loss: $\mathcal{L}^{\text{pix}} = \sum_t \|p_t^c - P_t v_t^c\|^2$, where $P_t$ is the projection matrix at frame $t$.

During training, foreground pixels are selected using the instance mask, and only those with valid CSE predictions and confidence scores above 0.5 are retained. For blurry frames with no valid CSE output, we reuse predictions from the previous frame.

\begin{figure}[t]
  \centering
\includegraphics[width=0.99\linewidth]{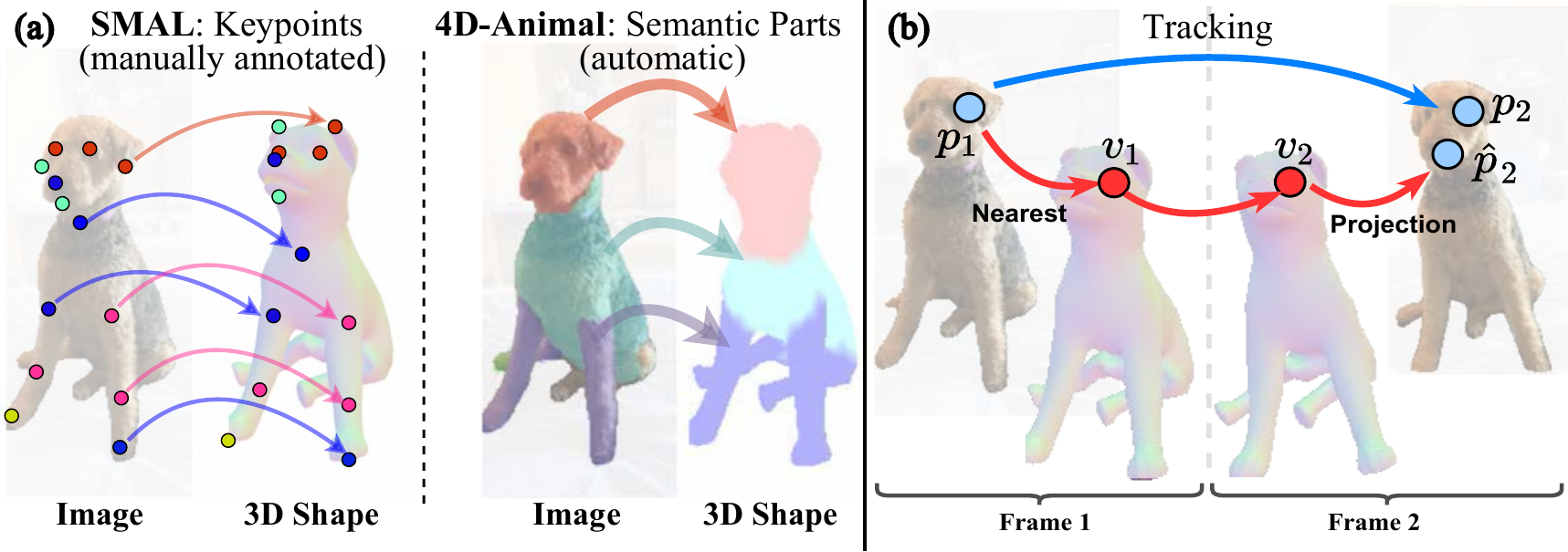}
  \caption{(a) \textbf{Left}: Keypoint alignment (manually annotated) . \textbf{Right}: Part-level alignment (automatic) using semantic masks divided into head, body, feet, and tail, which are mapped to corresponding SMAL mesh regions. (b) Temporal tracking alignment. A tracked 2D point $p_1$ in Frame 1 is matched to $p_2$ in Frame 2. The nearest vertex $v_1$ to $p_1$ is found, and its corresponding vertex $v_2$ is located via SMAL topology. The loss is computed between $p_2$ and the projection of $v_2$. 
  }
  \vspace{-15pt}
\label{fig:introduction}
\end{figure}

\subsubsection{Temporal tracking via SMAL topology}  
At the \emph{temporal level}, we use 2D tracking from BootsTAP~\cite{doersch2024bootstap}. BootsTAP is a model that can track any point on solid surfaces in a video. Specifically, We sample $N_t$ points uniformly within instance mask of Frame $t$ by randomly selecting pixel locations with equal probability, and track them in all frames of the video using BootsTAP. To ensure reliability, we filter any point whose tracked positions fall outside the instance mask in any frame, as this may indicate tracking drift or failure. 

Furthermore, since COP3D videos consist of 200 frames captured by a fly-around camera to ensure full-body coverage, we initialize point tracking from four key frames: 1, 51, 101, and 151. For each key frame, we sample $N_t = 500$ points and track them across the entire sequence. This results in a total of 2000 tracked points per video. After filtering out unreliable trajectories, approximately 1000 valid ones are typically retained.

Next, as shown in Fig.~\ref{fig:introduction}(b), for each tracked 2D point $p_1 \rightarrow p_2$ obtained from BootsTAP, we first project all mesh vertices from Frame 1 to the screen and identify the vertex $v_1$ closest to $p_1$. Given that the meshes $m_1$ and $m_2$ from Frames 1 and 2 share the same topology, we locate the corresponding vertex $v_2$ in $m_2$ by matching the index of $v_1$. Projecting $v_2$ to the screen yields the estimated location $\hat{p}_2$. This allows us to enforce temporal consistency by penalizing the discrepancy between the tracked point $p_2$ and the projected vertex $\hat{p}_2$.
Formally, for a set of 2D trajectories $p_t \rightarrow p_t'$ between frames $t$ and $t'$, we project the corresponding mesh vertices $v_t'$ to the screen using the projection matrix $P_{t'}$, and define the temporal consistency loss as: $\mathcal{L}^{\text{time}} = \sum_t \|p_t' - P_t' v_t'\|^2$.

\subsection{The novelty of 4D-Animal}

The primary motivation behind 4D-Animal arises from the limitations of existing model-based methods for animatable 3D animal reconstruction, which still rely on sparse semantic keypoints. Annotating such keypoints is labor-intensive, and keypoint detectors, often trained on limited animal datasets, suffer from unreliability. Even the most recent model-based methods~\cite{sabathier2024animal} that incorporate CSE~\cite{Neverova2020ContinuousSurfaceEmbeddings} still require some degree of keypoint supervision, which inherently limits their robustness, generalizability, and scalability.

In contrast, 4D-Animal departs from this paradigm by leveraging well-trained 2D vision models to eliminate the need for sparse keypoint annotations, enabling robust and animatable 3D reconstruction from video. Our contributions are twofold:
\begin{itemize}
     \item \textbf{Efficient feature mapping for SMAL fitting.} We propose a simple-yet-effective feature network that maps 2D features from pre-trained vision models to SMAL parameters. This design ensures high-quality feature input, improving both the efficiency and accuracy of the SMAL fitting process. Unlike prior work~\cite{sabathier2024animal}, which often relies on generic feature representation, our feature mapping is specifically optimized for animatable 3D animal reconstruction, balancing simplicity and performance.
     \item \textbf{Hierarchical alignment for robust reconstruction.} To further enhance the robustness of video-based reconstruction, we propose a hierarchical alignment strategy that operates across pixel-to-vertex correspondences, body part segmentations, and point trajectories. This strategy enables joint optimization of shape, pose, and motion in the absence of annotated keypoints, providing strong geometric and temporal cues derived solely from visual evidence.
\end{itemize}

Extensive experiments demonstrate that 4D-Animal outperforms both model-based and model-free baselines, achieving superior reconstruction quality without requiring sparse keypoint annotations. Additionally, the 3D assets generated by our method are beneficial for downstream tasks, highlighting its potential for scalable and generalizable animal modeling.

Finally, we emphasize that while 2D cues such as pixel-to-vertex correspondence~\cite{shtedritski2024shic,sommer2024unsupervised,de20244dpv}, part segmentation~\cite{liu2023lepard,li2020self,aygun2024saor}, and point tracking~\cite{yang2021viser,lei2024mosca,wang2024shape}, are well-established tools in computer vision and have seen widespread application in 3D/4D reconstruction, their systematic and task-driven integration into a keypoint-free 4D animal reconstruction pipeline is novel. Our method is distinguished by its careful integration of these components, specifically tailored to address the unique challenges of animatable animal shape and motion reconstruction.

\section{Additional details and results of experiments}

\subsection{Training procedure}
\label{sec:training}

\begin{figure}[ht]
  \centering
\includegraphics[width=0.69\linewidth]{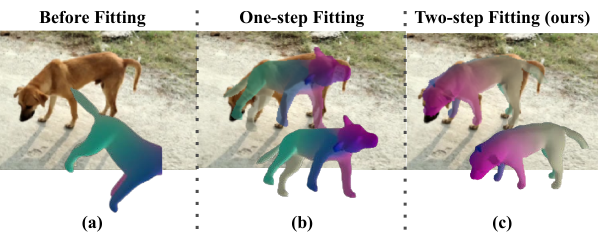}
\vspace{-5pt}
  \caption{(a) The initial pose of the mesh is often inaccurate. (b) In one-step fitting, the model tends to confuse the head and tail, focusing primarily on silhouette coverage. (c) Our two-step fitting strategy enables a more precise alignment, resulting in higher-quality reconstruction.
  }
  \vspace{-10pt}
\label{fig:training}
\end{figure}

As shown in Fig.\ref{fig:training} (a), the initial 3D pose is often inaccurate. Applying all loss terms uniformly from the beginning can lead to unstable optimization. For instance, as shown in Fig.\ref{fig:training} (b), the model may confuse semantic parts and prioritize silhouette coverage over precise alignment. To address this, we adopt a multi-stage fitting strategy.

In the early stage of training, we focus on stabilizing the global pose. We reduce the weight of the object-level alignment loss ($\lambda^{\text{obj}}$) and the temporal level loss ($\lambda^{\text{time}}$), and disable vertex offsets in the SMAL template. This encourages coarse alignment of the overall shape without introducing unstable local deformations or enforcing premature temporal constraints.

In the later stage, once the global pose has stabilized, we begin refining local geometry. Vertex offsets are enabled to allow detailed shape adjustment, especially along the silhouette. Meanwhile, the weights of pixel-level and part-level losses ($\lambda^{\text{pixel}}, \lambda^{\text{part}}$) are gradually reduced to avoid overfitting to noisy or ambiguous regions. In contrast, the temporal consistency loss ($\lambda^{\text{time}}$) is progressively increased to enforce smooth motion and coherent alignment across frames. As shown in Fig.~\ref{fig:training} (c), this leads to more accurate and temporally consistent reconstructions.

In our experiments, training runs for 10,000 epochs. The loss weights are scheduled as follows:

\begin{itemize}
     \item $\lambda^{\text{obj}}$: 1, 100, 500, 800 at milestones 300, 1000, 6000.
     \item $\lambda^{\text{part}}$: 5e-4, 5e-8 at milestones 300.
     \item $\lambda^{\text{pixel}}$: 5, 1e-1, 1e-2 at milestones 1000, 6000.
     \item $\lambda^{\text{time}}$: 5e-4, 5e-2 at milestones 300.
\end{itemize}

As shown in Fig.~\ref{fig:efficiency}, although our performance is initially lower than that of Avatars during the early fitting stage, it quickly surpasses Avatars and ultimately achieves the target metric values with significantly higher efficiency.

\begin{figure}[ht]
  \centering
\includegraphics[width=0.60\linewidth]{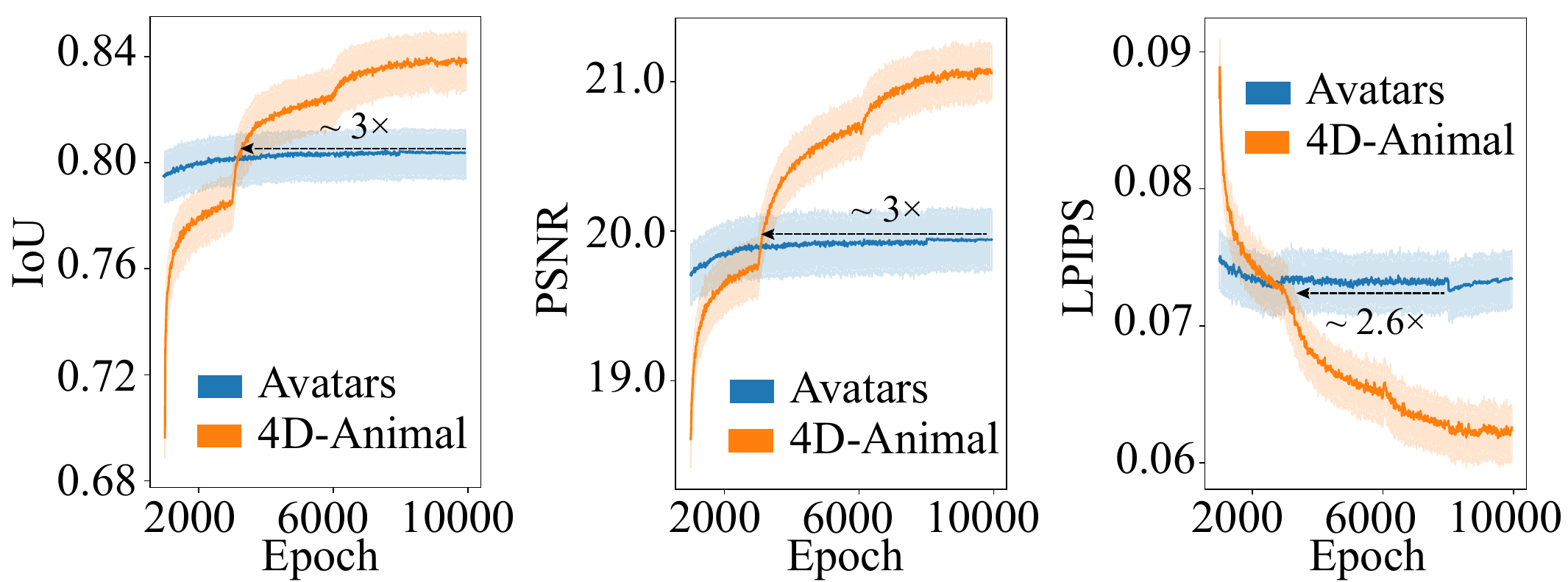}
\vspace{-5pt}
  \caption{Comparison of training efficiency between 4D-Animal and Avatars. }
  \vspace{-10pt}
\label{fig:efficiency}
\end{figure}

\subsection{Implementation details}
We evaluate all models on COP3D~\cite{sinha2023common}, a publicly available dataset featuring fly-around videos of pets, with annotated camera parameters and object masks. Following~\cite{sabathier2024animal}, we select a subset of 50 dog videos capturing diverse poses, motions, and textures. Each test video contains 200 frames, which we split into the training
and test sets by considering contiguous blocks of 15 frames as train, interleaved by blocks of 5 frames as test. The videos we use are the same as those used in Avatars~\cite{sabathier2024animal}.
We also use casual videos with depth maps of dogs from TracksTo4D~\cite{kasten2024fast} to evaluate the reconstructed structure.

We employ the Adam optimizer and apply a learning rate decay to each parameter group by a factor of $\gamma$ when the epoch count reaches predefined milestones. For texture reconstruction, $\gamma$ is set to 0.5 with milestones at [9000, 9500]. The number of epochs is set to 10000. The batch size is set as 32. The training and evaluation of all models is conducted using the RTX 6000 GPU (48G).

\subsection{Introduction of baselines}

We compare 4D-Animal against the state-of-the-art model-based counterparts, including Avatars~\cite{sabathier2024animal}, BARC~\cite{rueegg2022barc}, and BITE~\cite{ruegg2023bite}, as well as the model-free approach RAC~\cite{yang2023reconstructing}. 
\begin{itemize}
     \item BARC~\cite{rueegg2022barc} recovers the 3D shape and pose of dogs from a single image by leveraging breed information. It modifies the SMAL animal model to better represent dog shapes and addresses the challenge of limited 3D training data by incorporating breed-specific losses during training. 
     \item BITE~\cite{ruegg2023bite} improves 3D dog pose estimation by introducing a dog-specific model (D-SMAL) and leveraging ground contact constraints to refine poses. A neural network enhances initial predictions by integrating contact information, enabling realistic reconstructions of complex postures. 
     \item RAC~\cite{yang2023reconstructing} reconstructs animatable 3D models of object categories like humans, cats, and dogs from monocular videos by disentangling morphology (shape variations) and articulation (motion over time). It achieves this by learning a category-level skeleton, using latent space regularization to maintain instance-specific details, and leveraging 3D background models for better segmentation.
     \item Avatars~\cite{sabathier2024animal} reconstructs animatable 3D dog models from monocular videos by jointly optimizing shape, pose, and texture in a canonical space. It improves model-based reconstruction using Continuous Surface Embeddings for dense keypoint supervision and introduces a duplex-mesh implicit texture model for realistic appearance. 
\end{itemize}

\subsection{Quantitative results of Avatars}

To ensure a fair comparison, we re-run the official codebase of Avatars~\cite{ruegg2023bite} using the same experimental setup as our method. The numerical results reported in Table~1 are based on these re-runs. We observe that our re-run results are slightly lower than those reported in the original paper. This gap may come from several factors, including:

\begin{itemize}
\item \textbf{Incomplete configuration details}: Some hyperparameters, such as loss weights or learning rate schedules, may not be fully specified in the released codebase.
\item \textbf{Differences in environment}: Variations in hardware, software versions, or random seeds can lead to slight performance fluctuations.
\end{itemize}

Despite the discrepancy, we use our re-run results for Avatars to maintain fairness, since all other re-run baselines are evaluated under the same experimental setup.

\subsection{Quantitative results of Avatars fitting without sparse keypoints}

Although Avatars~\cite{sabathier2024animal} incorporate Continuous Surface Embeddings (CSE)\cite{Neverova2020ContinuousSurfaceEmbeddings}, which provide dense image-to-mesh constraints that are stronger than traditional sparse semantic keypoint supervision, their final model still relies on sparse keypoints. As noted in \cite{sabathier2024animal}, keypoints are particularly helpful in handling challenging regions such as thin structures, where CSE predictions may be less reliable.

To assess the impact of sparse keypoint supervision, we compare the performance of Avatars and 4D-Animal with and without keypoints, as shown in Table~\ref{tab:keypoints}. Both methods show performance improvements when trained with keypoints. However, the relative gain for Avatars is more pronounced, indicating that its performance benefits more directly from keypoint annotations. In contrast, 4D-Animal maintains competitive accuracy even without keypoints, suggesting that its hierarchical alignment, which integrates silhouette, part-level, pixel-level, and temporal cues, effectively maintains reconstruction accuracy without relying on sparse keypoints.

These findings demonstrate that 4D-Animal offers greater robustness and scalability compared to approaches that rely on keypoint annotations, particularly in large-scale or in-the-wild deployment scenarios where keypoint labeling is costly, labor-intensive, or unavailable.

\begin{table}[t]
  \caption{Quantitative evaluation of pose estimation measured by IoU, IoUw5 on 50 sequences from COP3D~\cite{sinha2023common}. SK: training with sparse keypoints. We also report the percentage improvement achieved by methods trained with SK over those trained without it.}
  \centering
  \renewcommand{\arraystretch}{1.3}
  \resizebox*{0.39\linewidth}{!}{
    \begin{tabular}{ll|ccccc}
    \toprule
    Models & SK & IoU ↑  & IoUw5 ↑ \\
    \midrule
    Avatars~\cite{sabathier2024animal} &  & 0.78  & 0.64  \\
    Avatars~\cite{sabathier2024animal} & \checkmark & 0.81  & 0.66  \\
    \rowcolor[rgb]{ .851,  .851,  .851} Improv. &  & 3.85\%  & 3.13\% \\
    \midrule
    4D-Animal &  & 0.84 & 0.71 \\
    4D-Animal & \checkmark & 0.85 & 0.72 \\
    \rowcolor[rgb]{ .851,  .851,  .851} Improv. &  & 1.19\% & 1.41\% \\
    \bottomrule
    \end{tabular}%
  }
  \label{tab:keypoints}%
\vspace{-10pt}
\end{table}%

\subsection{Qualitative comparison of hierarchical alignment}

The quantitative impact of hierarchical alignment is presented in Section 6 of the main text. Here, we provide a qualitative analysis based on the ablation study to further understand the contribution of each alignment component.

\begin{figure*}[h]
  \centering
\includegraphics[width=0.8\linewidth]{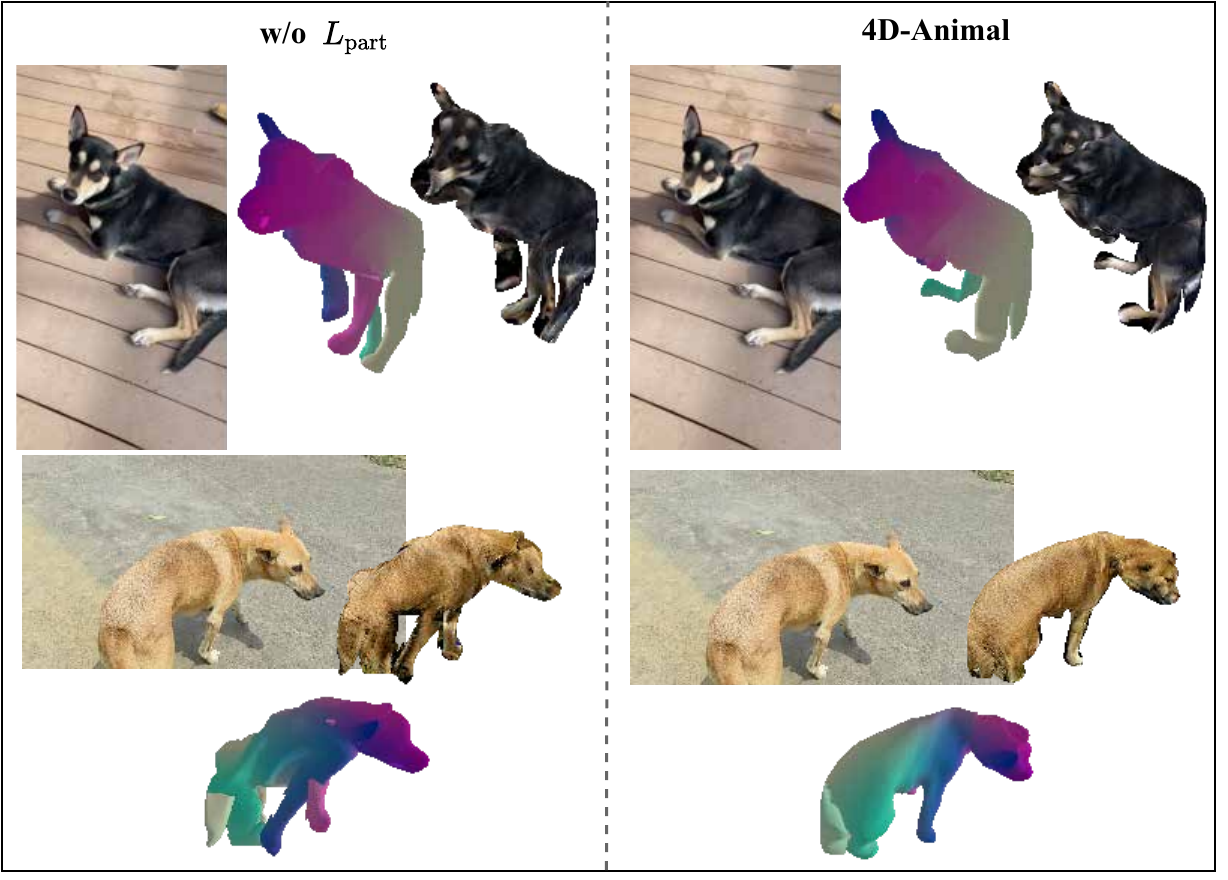}
  \caption{Qualitative ablation of part-level alignment $L_{\text{part}}$. }
\label{fig:qualitative_hierarchical_part}
\end{figure*}

First, as shown in Fig.~\ref{fig:qualitative_hierarchical_part}, removing the part-level loss $L_{\text{part}}$ leads to failures in cases involving fast camera or animal motion. In such scenarios, temporal tracking and pixel-to-vertex correspondences are often unreliable, making semantic part masks an essential source of guidance.

\begin{figure*}[h]
  \centering
\includegraphics[width=0.8\linewidth]{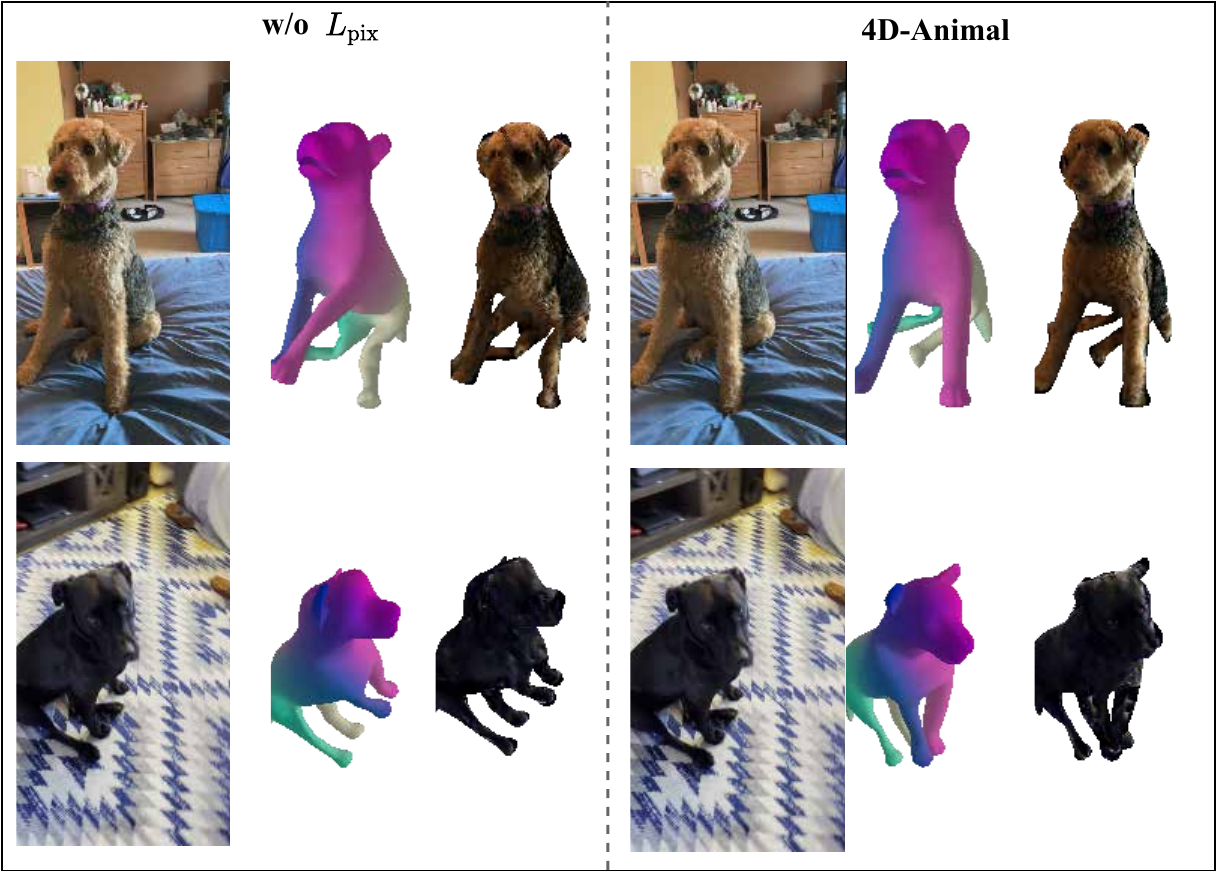}
  \caption{Qualitative ablation of pixel-level alignment $L_{\text{pix}}$. }
\label{fig:qualitative_hierarchical_pix}
\end{figure*}

Second, Fig.~\ref{fig:qualitative_hierarchical_pix} shows that while the 4D-Animal without pixel-to-vertex supervision ($L_{\text{pix}}$) can still recover the overall pose, it struggles to capture finer details such as the accurate articulation of the head and legs.

\begin{figure*}[h]
  \centering
\includegraphics[width=\linewidth]{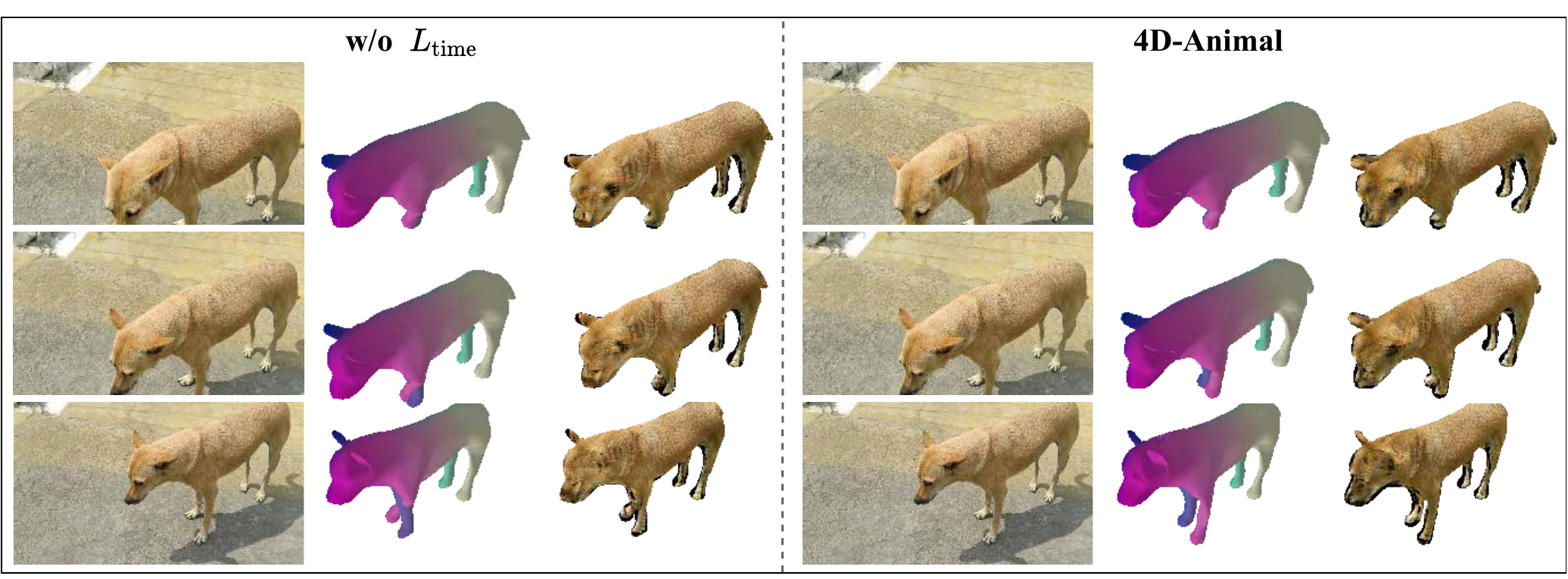}
  \caption{Qualitative ablation of temporal-level alignment $L_{\text{time}}$.}
\label{fig:qualitative_hierarchical_time}
\end{figure*}

Finally, as shown in Fig.~\ref{fig:qualitative_hierarchical_time}, the temporal consistency loss $L_{\text{tracking}}$ helps maintain coherent leg behavior over time. Without $L_{\text{tracking}}$, 4D-Animal may reconstruct inconsistent poses across frames. For instance, reconstructing crossed legs in later frames that contradict the pose observed earlier. In contrast, incorporating $L_{\text{tracking}}$ ensures temporal consistency, preventing such contradictions and leading to accurate and stable reconstructions.

In summary, each component of the hierarchical alignment plays a crucial role in handling different challenges, and their combination improves the robustness of 4D-Animal in complex video scenarios.

\subsection{Novel view of animatable 3D assets}
\begin{figure}[ht]
  \centering
\includegraphics[width=0.79\linewidth]{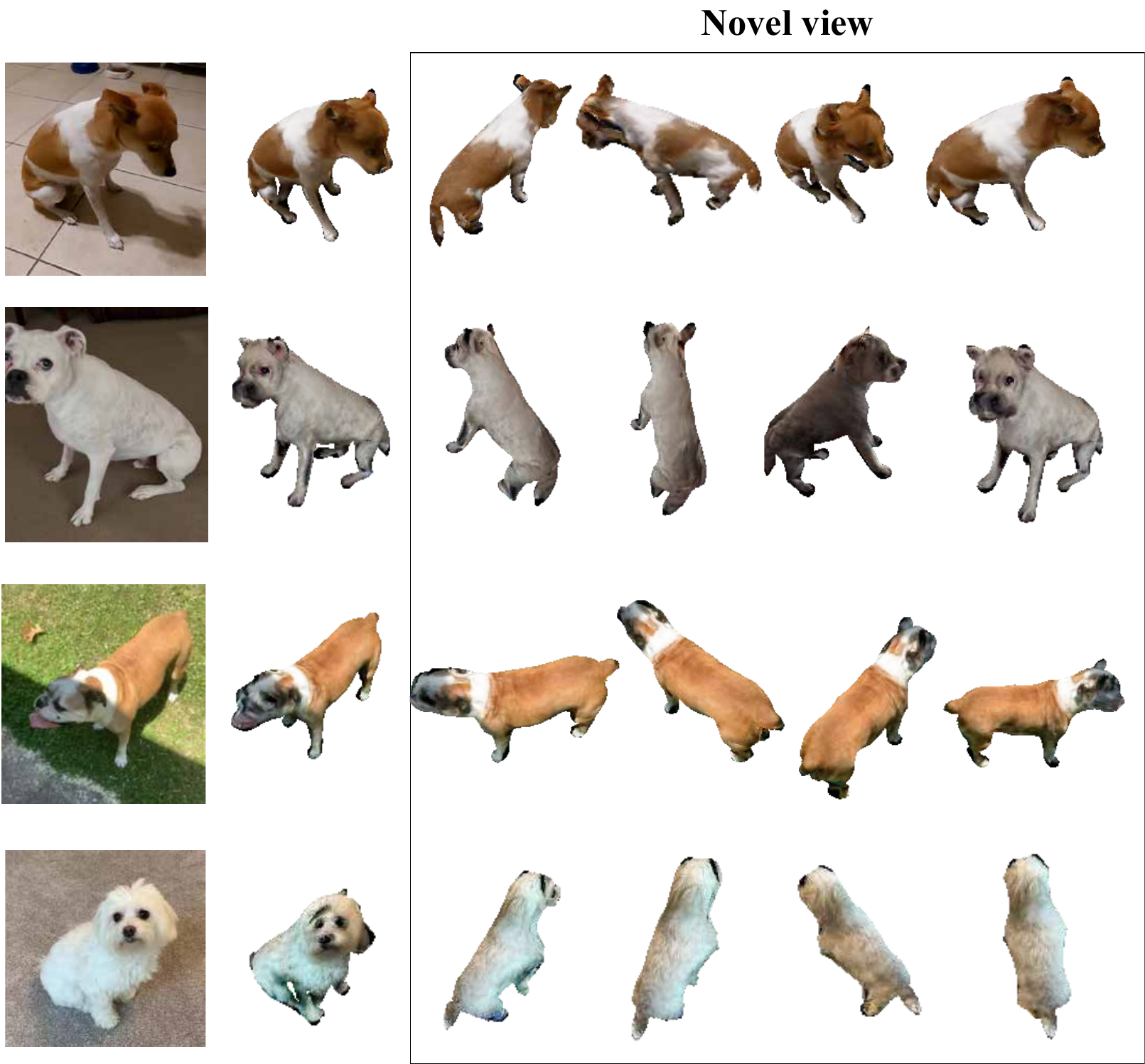}
  \caption{Novel views of animatable 3D assets reconstructed using 4D-Animal.}
\label{fig:novel_view}
\end{figure}

We provide additional materials showcasing novel views of the animatable 3D assets reconstructed using our 4D-Animal, as shown in Fig.~\ref{fig:novel_view}, to offer a comprehensive evaluation of 4D-Animal's performance. These views highlight 4D-Animal's ability to generate consistent reconstructions from previously unseen perspectives, further demonstrating the robustness and generalization capability of 4D-Animal.

\clearpage
\subsection{Reconstruction on videos of other animal types}

\textbf{Why use a dog-specific SMAL model?} 
In our experiments, we implement 4D-Animal using the state-of-the-art dog-specific SMAL template~\cite{ruegg2023bite} to ensure a focused and comprehensive evaluation, minimizing external factors such as template limitations. However, we acknowledge the importance of generalizing our method to other animal categories and species, and we clarify that the core of our method is indeed designed to be broadly applicable.

\textbf{Why can our method be generalized to other animal categories?} 
The key components of our pipeline, including hierarchical alignment cues such as object masks, semantic part masks, pixel-to-vertex correspondences, and motion tracking, are not tied to any specific animal species. These cues can be applied across different quadruped species without significant modification. Our method relies on dense feature networks that align 2D representations to SMAL parameters, and since these features are general and not category-dependent, they are expected to work similarly across other quadrupeds. 

\begin{figure}[bh]
  \centering
\includegraphics[width=\linewidth]{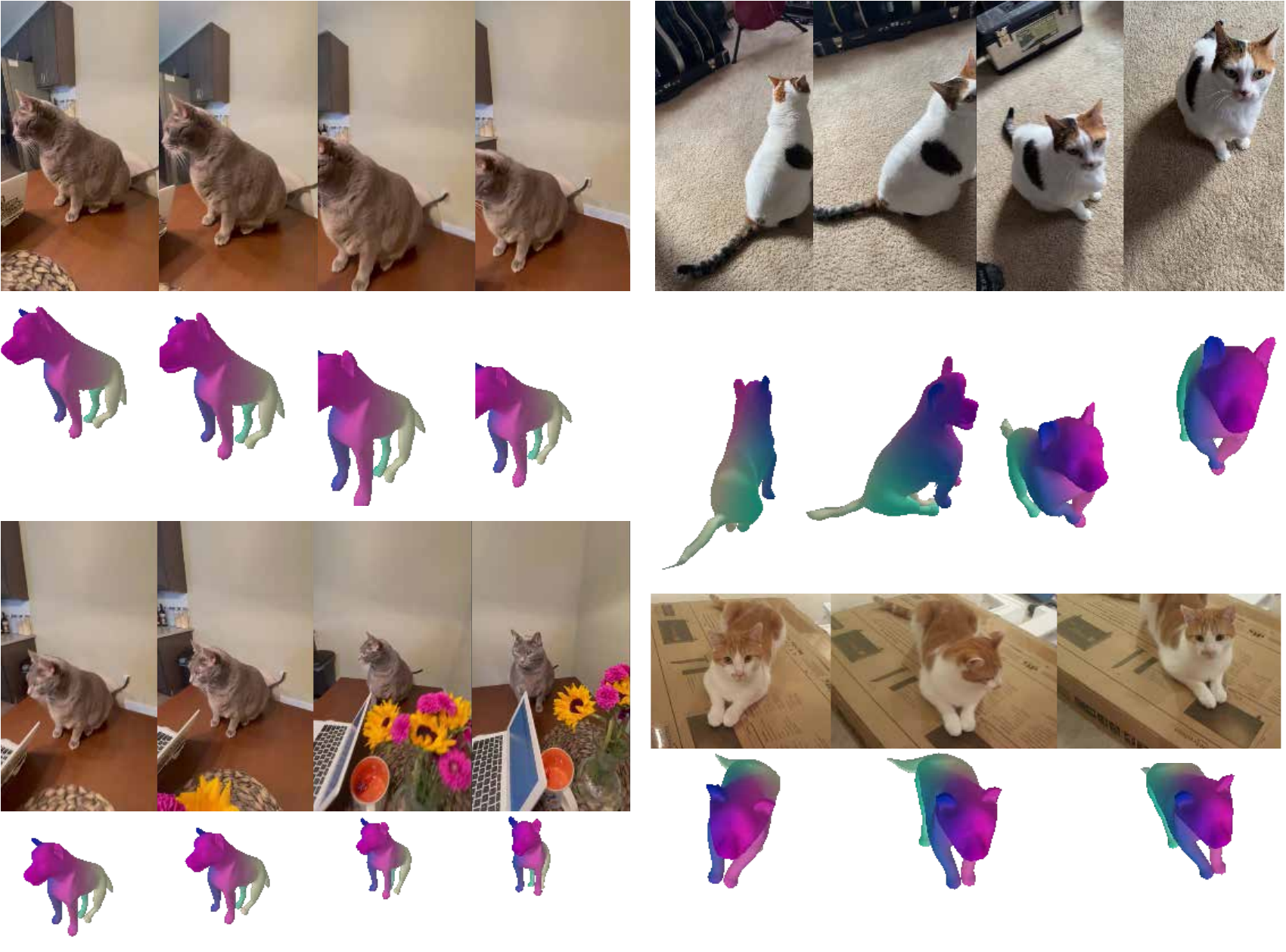}
  \caption{Qualitative results of 4D-Animal on cat videos from COP3D. Despite using a dog-specific template, the method achieves reasonable mesh fittings, demonstrating its potential generalization to other animal categories.}
\label{fig:other_animal}
\end{figure}

\textbf{How does our method perform on other animal categories?} 
We conduct preliminary tests of our method on cat videos from the COP3D dataset, as shown in Fig.\ref{fig:other_animal}. Despite using a dog-specific template, 4D-Animal still reconstructs reasonable mesh fittings to the 2D images. Although some details such as the head are less accurate, the overall results remain promising.

Although our experiments focus on dogs as a proof-of-concept, the proposed approach is inherently generalizable. We believe it holds strong potential for extension to a wide range of quadruped animals.

\subsection{Additional comparison of reconstruction quality and temporal consistency}

\begin{figure*}[ht]
  \centering
\includegraphics[width=0.75\linewidth]{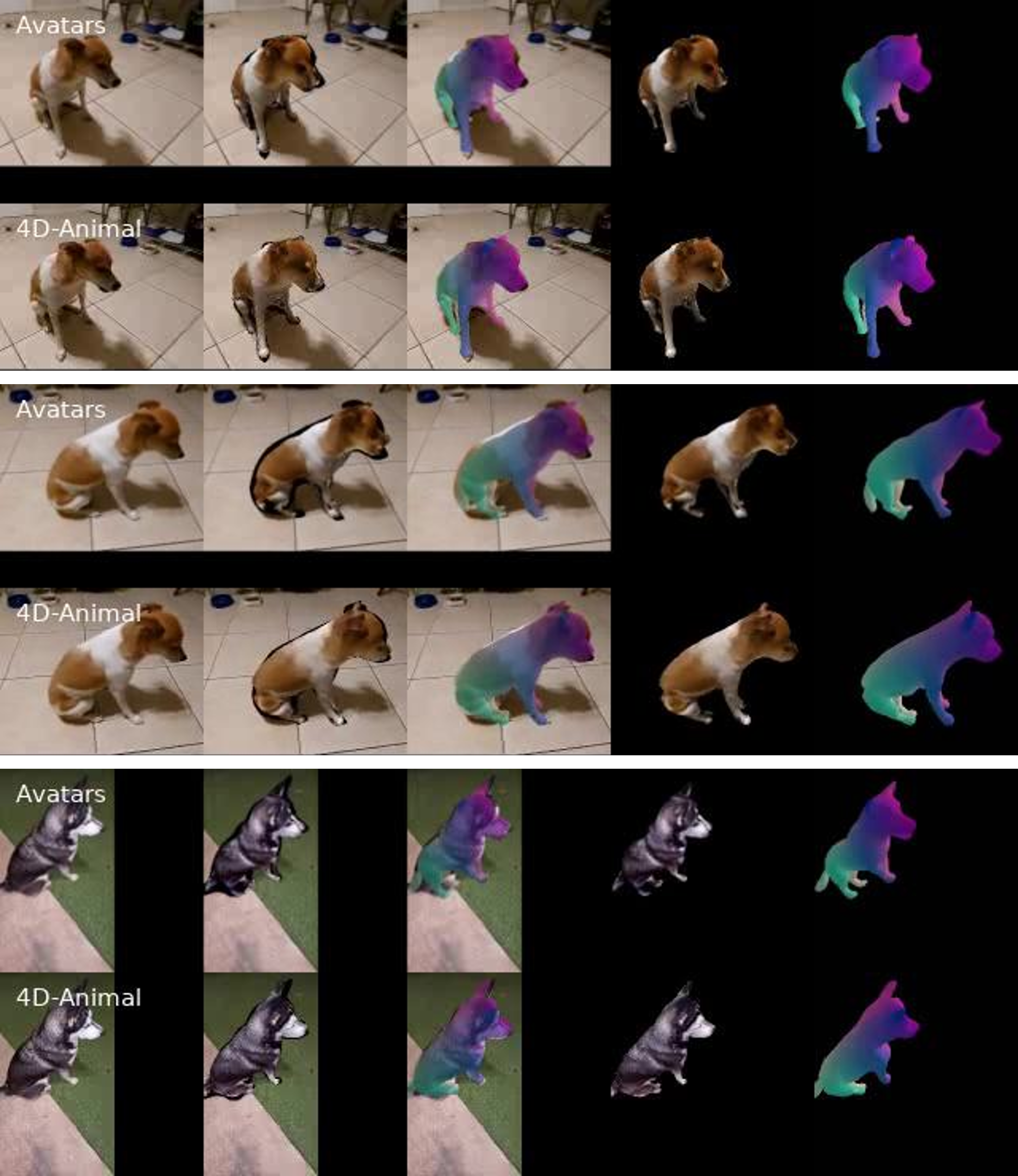}
  \caption{Qualitative comparison. We compare our 4D-Animal with the state-of-the-art model Avatars~\cite{sabathier2024animal} by selecting images from videos. The \textbf{first column} shows the original images, the \textbf{second column} presents the reconstructed appearance overlaid on the original image, and the \textbf{third column} displays the reconstructed 3D shape overlaid on the original image. The \textbf{fourth} and \textbf{fifth columns} show the standalone rendered appearance and 3D shape, respectively.}
\label{fig:additional_visualization_1}
\end{figure*}

\begin{figure*}[t]
  \centering
\includegraphics[width=0.75\linewidth]{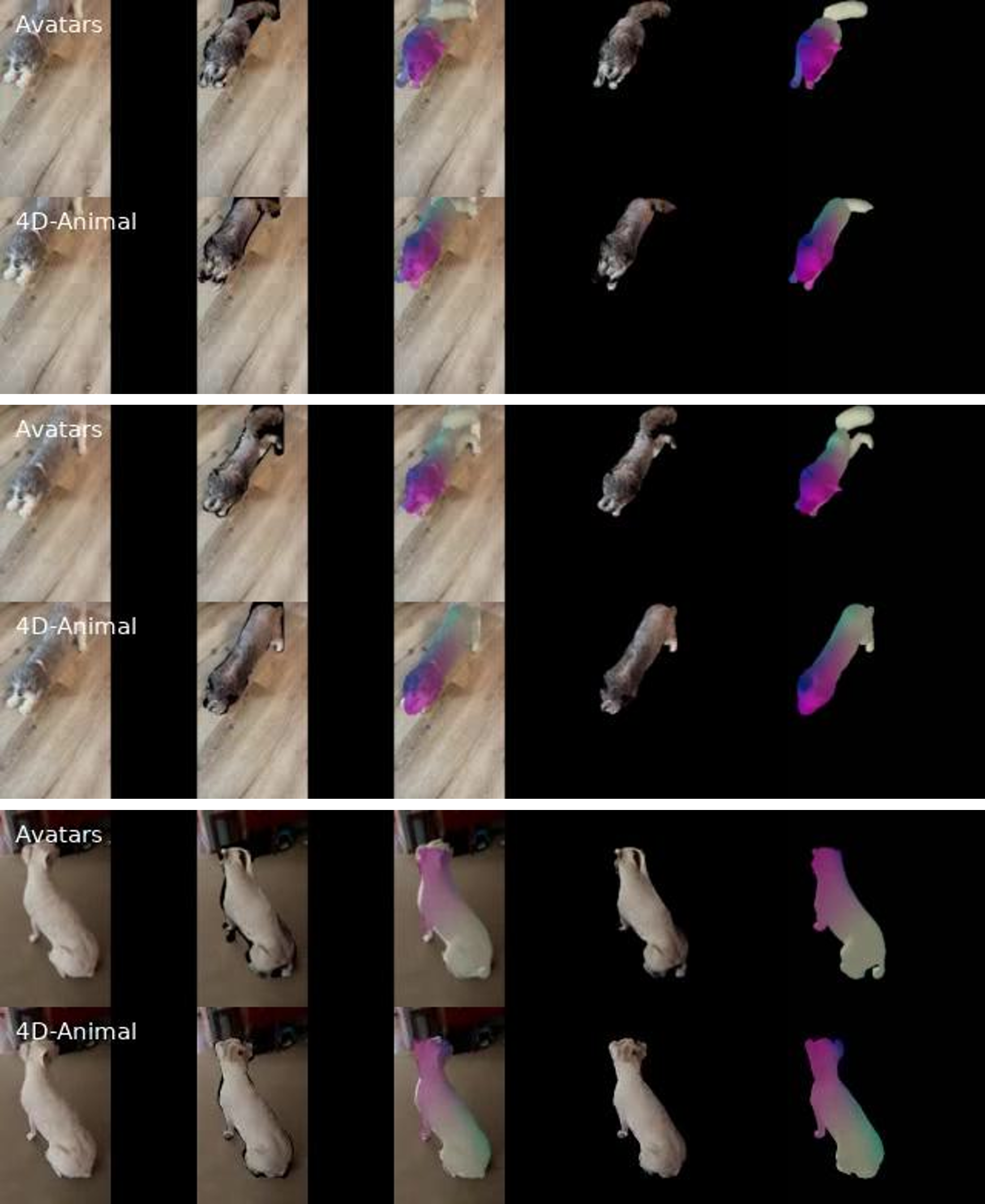}
  \caption{Qualitative comparison. We compare our 4D-Animal with the state-of-the-art model Avatars~\cite{sabathier2024animal} by selecting images from videos. The \textbf{first column} shows the original images, the \textbf{second column} presents the reconstructed appearance overlaid on the original image, and the \textbf{third column} displays the reconstructed 3D shape overlaid on the original image. The \textbf{fourth} and \textbf{fifth columns} show the standalone rendered appearance and 3D shape, respectively.}
\label{fig:additional_visualization_2}
\end{figure*}

\begin{figure*}[t]
  \centering
\includegraphics[width=0.75\linewidth]{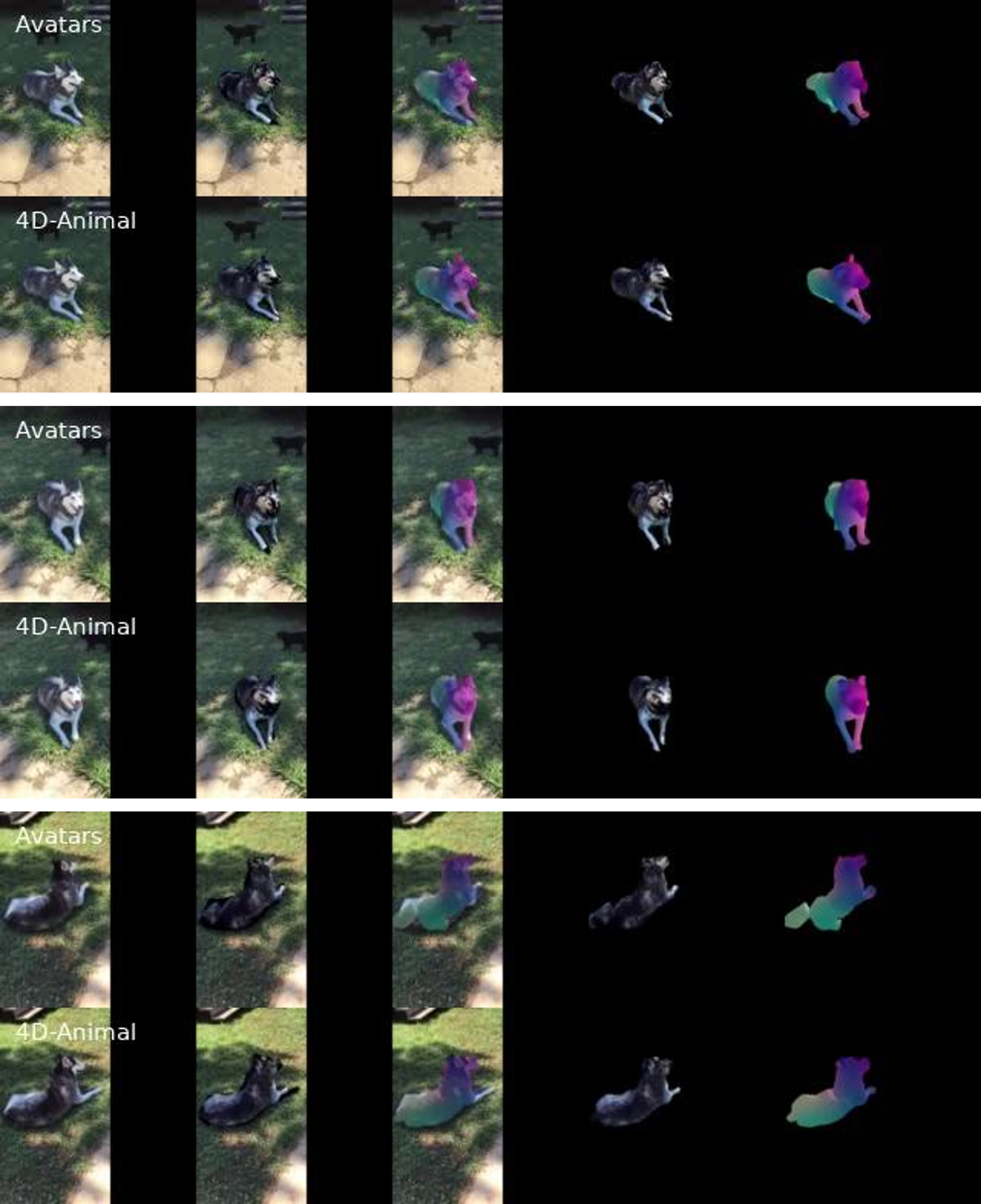}
  \caption{Qualitative comparison. We compare our 4D-Animal with the state-of-the-art model Avatars~\cite{sabathier2024animal} by selecting images from videos. The \textbf{first column} shows the original images, the \textbf{second column} presents the reconstructed appearance overlaid on the original image, and the \textbf{third column} displays the reconstructed 3D shape overlaid on the original image. The \textbf{fourth} and \textbf{fifth columns} show the standalone rendered appearance and 3D shape, respectively.}
\label{fig:additional_visualization_3}
\end{figure*}

\begin{figure*}[t]
  \centering
\includegraphics[width=0.75\linewidth]{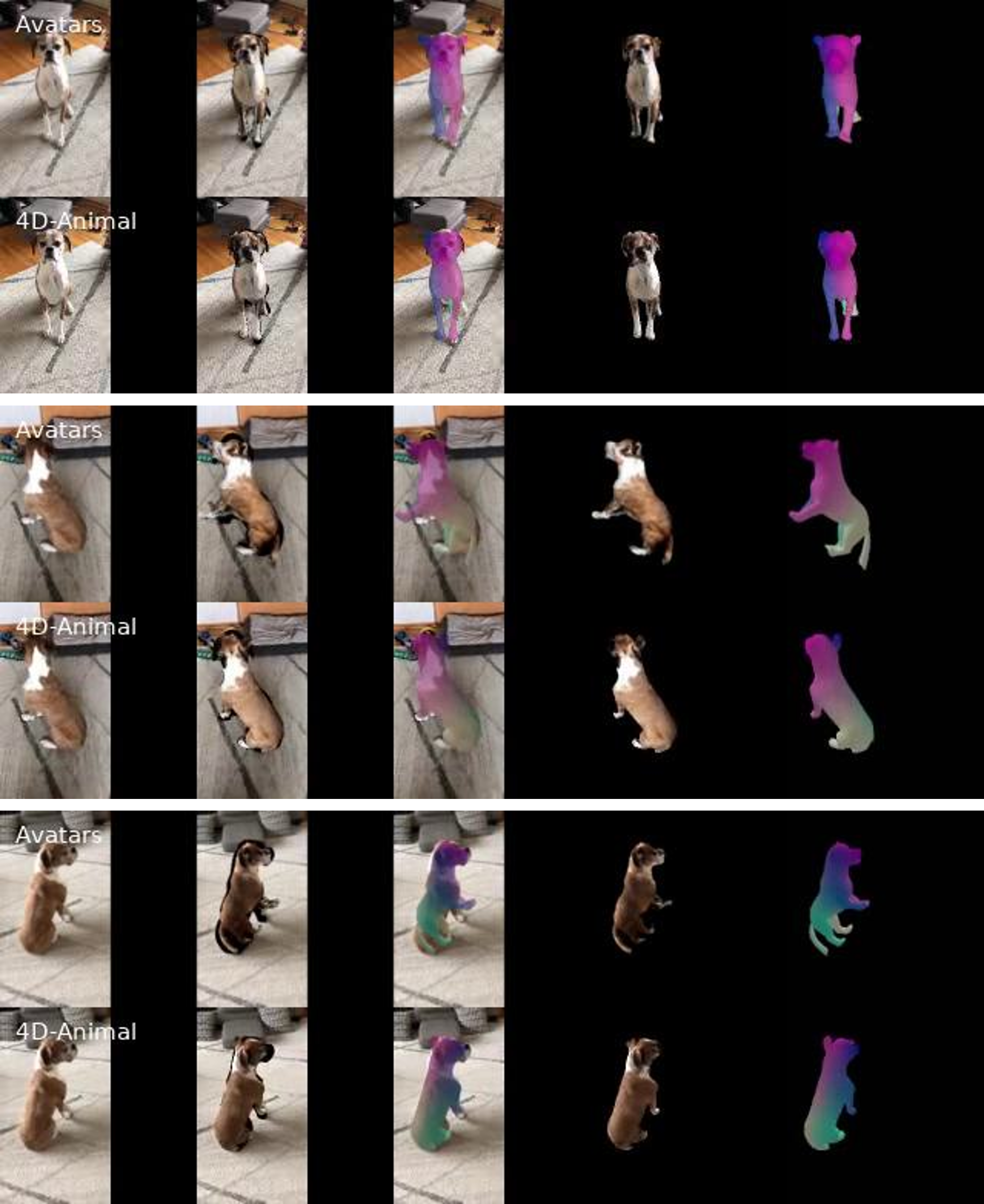}
  \caption{Qualitative comparison. We compare our 4D-Animal with the state-of-the-art model Avatars~\cite{sabathier2024animal} by selecting images from videos. The \textbf{first column} shows the original images, the \textbf{second column} presents the reconstructed appearance overlaid on the original image, and the \textbf{third column} displays the reconstructed 3D shape overlaid on the original image. The \textbf{fourth} and \textbf{fifth columns} show the standalone rendered appearance and 3D shape, respectively.}
\label{fig:additional_visualization_4}
\end{figure*}

\begin{figure*}[t]
  \centering
\includegraphics[width=0.75\linewidth]{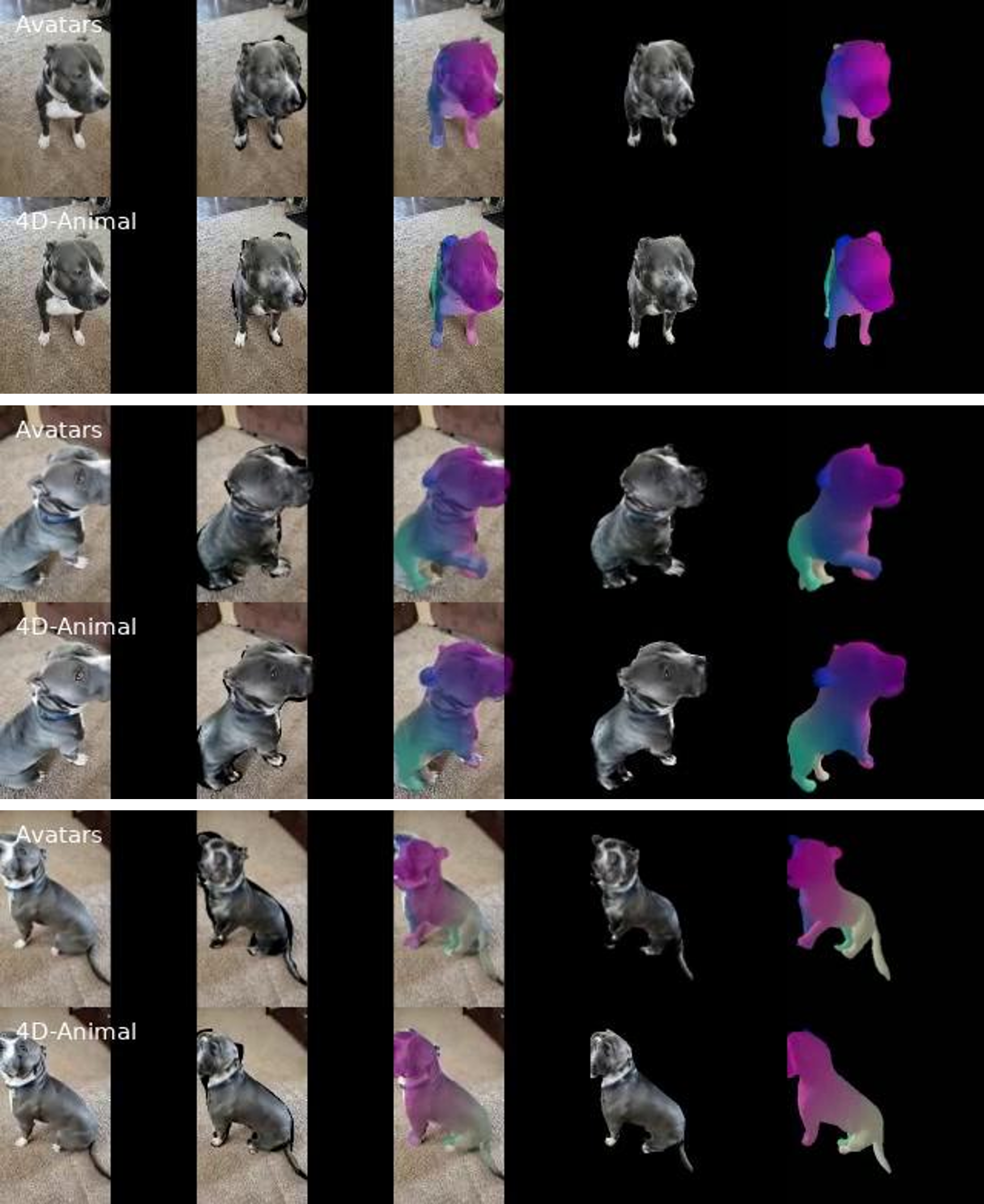}
  \caption{Qualitative comparison. We compare our 4D-Animal with the state-of-the-art model Avatars~\cite{sabathier2024animal} by selecting images from videos. The \textbf{first column} shows the original images, the \textbf{second column} presents the reconstructed appearance overlaid on the original image, and the \textbf{third column} displays the reconstructed 3D shape overlaid on the original image. The \textbf{fourth} and \textbf{fifth columns} show the standalone rendered appearance and 3D shape, respectively.}
\label{fig:additional_visualization_5}
\end{figure*}

We provide further qualitative comparisons to evaluate the reconstruction quality of our 4D-Animal method against existing methods as shown in Fig.~\ref{fig:additional_visualization_1}, \ref{fig:additional_visualization_2}, \ref{fig:additional_visualization_3}, \ref{fig:additional_visualization_4}, \ref{fig:additional_visualization_5}. For a more detailed visualization and an evaluation of temporal consistency, please refer to the \textbf{supplementary videos}.

\clearpage

\section{Analysis of state-of-the-art image-to-3D models}

\begin{figure*}[ht]
  \centering
\includegraphics[width=0.99\linewidth]{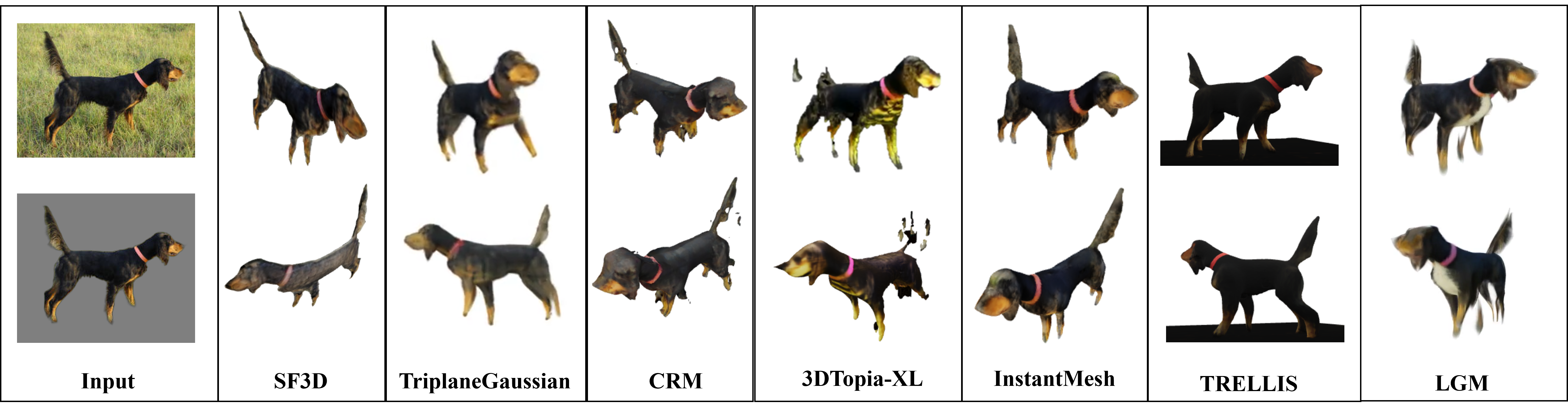}
  \caption{The performance of image-to-3D models SF3D~\cite{boss2024sf3d}, TriplaneGaussian~\cite{zou2024triplane}, CRM~\cite{wang2024crm}, 3DTopoa-XL~\cite{chen20243dtopia}, InstantMesh~\cite{xu2024instantmesh}, TRELLIS~\cite{xiang2024structured}, LGM~\cite{tang2024lgm} on animal images. }
\label{fig:image_to_3d_models}
\end{figure*}

As we have mentioned in Sec. 6 (main text), we observe that current 3D generative models trained on synthetic datasets often fail when processing real-world animal images. Here, we assess the performance of a wider variety of 3D models beyond LGM~\cite{tang2024lgm} as shown in Fig~\ref{fig:image_to_3d_models}, such as SF3D~\cite{boss2024sf3d}, TriplaneGaussian~\cite{zou2024triplane}, CRM~\cite{wang2024crm}, 3DTopoa-XL~\cite{chen20243dtopia}, InstantMesh~\cite{xu2024instantmesh} and TRELLIS~\cite{xiang2024structured}. There is still a gap between such models and model-based methods, but we can take full advantage of both. Fine-tuning existing large models using 4D-Animal reconstruction results of real-world casual videos could be very efficient and does not limit the representation of the models. This highlights that our work could form an essential contribution to the image-to-3D community, complementing existing efforts.

\end{document}